%% file: main.tex
\newcommand\blfootnote[1]{%
  \begingroup
  \renewcommand\thefootnote{}\footnote{#1}%
  \addtocounter{footnote}{-1}%
  \endgroup
}
\PassOptionsToPackage{dvipsnames, table}{xcolor}
\PassOptionsToPackage{hidelinks}{hyperref}

\documentclass{bmvc2k}

\usepackage{ifthen}
\newboolean{preprint}
\setboolean{preprint}{true}

\usepackage[utf8]{inputenc}
\usepackage{amsmath, amsthm, amsfonts, graphicx, amssymb, tabularx,subcaption}
\usepackage{multirow,bigdelim,bm,dsfont,mathtools}
\usepackage{adjustbox}
\usepackage{pgfplots, xspace, float,bm, tikz}
\usepackage[title]{appendix}
\usepackage{etoolbox, booktabs}
\usepackage{todonotes}
\usepackage{wrapfig}
\usepackage[capitalise,noabbrev]{cleveref}
\usepackage{algorithm,listings}
\usepackage[noend]{algpseudocode}
\usepackage[utf8]{inputenc}
\usepackage[frozencache,cachedir=.]{minted}
\usepackage[hashEnumerators,smartEllipses,hybrid,cacheDir=/tmp]{markdown}

\def\eg{\emph{e.g}\bmvaOneDot}

\def\ie{\emph{i.e}\bmvaOneDot}

\def\Paumer{\textsc{Paumer}}
\def\paumer{\textsc{paumer}}

\def\R{\ensuremath{\mathbb{R}}}
\def\loss{\ensuremath{\mathcal{L}}}
\def\pauseprop{\ensuremath{\tau}}
\def\dvit{DynamicViT\xspace}
\def\loss{\ensuremath{\mathcal{L}}}
\LetLtxMacro{\ORIcitep}{\citep}
\DeclareRobustCommand{\citep}{\leavevmode\unskip~\ORIcitep}

\LetLtxMacro{\ORIcitet}{\citet}
\DeclareRobustCommand{\citet}{\leavevmode\unskip~\ORIcitet}

\hypersetup{colorlinks=false,citecolor=MidnightBlue,linkcolor=red,urlcolor=brown,pdfborder={0 0 0}}
\pgfplotsset{compat=newest}

\ifthenelse{\boolean{preprint}}{}{\bmvcreviewcopy{737}}
\usetikzlibrary{positioning,decorations.markings, }

\AtBeginEnvironment{appendices}{\crefalias{section}{appendix}}

\addauthor{Evann Courdier$^\ast$}{evann.courdier@idiap.ch}{1,2}
\addauthor{Prabhu Teja S$^\ast$}{prabhu.teja@idiap.ch}{1,2}
\addauthor{Fran\c{c}ois Fleuret}{francois.fleuret@unige.ch}{1,2,3}

\addinstitution{
  Idiap Research Institute,\\Switzerland.
}
\addinstitution{
  EPFL, Switzerland.
}

\addinstitution{
  University of Geneva, \\Switzerland.
}

\runninghead{Courdier, Teja, Fleuret}{\Paumer\xspace for Semantic Segmentation}
\title{\Paumer: Patch Pausing Transformer for Semantic Segmentation}

\setuptodonotes{inline}

\begin{document}
\maketitle
\ifthenelse{\boolean{preprint}}{\blfootnote{$^\ast$ Equal Contribution.}}{}

\begin{abstract}
    We study the problem of improving the efficiency of segmentation transformers by using disparate amounts of computation
    for different parts of the image. Our method, \Paumer, accomplishes this by pausing computation for patches that are
    deemed to not need any more computation before the final decoder. We use the entropy of predictions computed from
    intermediate activations as the pausing criterion, and find this aligns well with semantics of the
    image. Our method has a unique advantage that a single network trained with the proposed strategy can be effortlessly
    adapted at inference to various run-time requirements by modulating its pausing parameters. On two standard segmentation
    datasets, Cityscapes and ADE20K, we show that our method operates with about a $50\%$ higher throughput with an mIoU
    drop of about $0.65\%$ and $4.6\%$ respectively.
\end{abstract}

\input{intro}
\input{proposed_method}
\input{previous_work}
\input{experiments}

\input{discussion}
\ifthenelse{\boolean{preprint}}{{\noindent
\paragraph{Acknowledgements}
Evann Courdier and Prabhu Teja are supported by the ``Swiss Center for Drones and Robotics - SCDR'' of the Swiss
Department of Defence, Civil Protection and Sport via armasuisse S+T under project No 050-38.}
}{}

\bibliography{main}

\clearpage
\renewcommand{\appendixpagename}{Supplementary Sections}
\begin{appendices}
    \appendixpage
    \input{appendix}
\end{appendices}

\end{document}

%% file: intro.tex
\section{Introduction}\label{sec:introduction}

Vision transformers\citep{dosovitskiy2020vit,steiner2021augreg} (ViT) have recently demonstrated very strong performance on large scale
image classification tasks. These networks break the images into a collection of patches (or tokens, interchangeably) and
progressively refine their representation by processing them through a series of residual self-attention
layers\citep{vaswani2017attention}. While their genesis was for image classification, recent methods have adapted
transformer architectures to various computer vision tasks\citep{salman2021transformerssurvey,
wang2021pyramid,xiangxiang2021twins}, and specifically to semantic
segmentation\citep{zheng2021rethinking,strudel2021segmenter,xie2021segformer}.

While these large transformer architectures have lead the progress on the accuracy front, there have been several
efforts to make them more efficient to be able to process more data, and faster\citep{tay2020efficient}. One way to
achieve this is to reduce the number of processed patches. Some works use multiscale
approaches\citep{wang2021pyramid,xie2021segformer} that gradually reduce the number of patches as the processing
progresses. Another option has been to drop the patches that are not informative to the classification
task\citep{yin2022avit,pan2021IARED2IR,marin2022token,rao2021dynamicvit,liang2022evit}.
For example, it is possible to classify an image as that of a dog with only the patches that belong to the
dog, while refraining from processing the rest of the patches.

\begin{figure}
    \centering
    \resizebox{\textwidth}{!}{\input{figures/example1}}
    \vspace{2ex}
    \caption{Illustration of our proposed method. Our method progressively stops processing patches after they reach a low
        enough prediction entropy. First column (a) shows the input image. Second column (b.1) shows the patches that are
        stopped from being processed after the third transformer layer in \textcolor{Aquamarine}{green}. Third column shows
        additional patches that are paused after the fifth layer in \textcolor{Salmon}{pink}. In the bottom row, (b.2) and
        (c.2), we show the entropy computed from the auxiliary decoders that is used to decide which patches to
        pause (\cref{sec:whyentropy}). It is apparent that the network automatically pauses easy parts of the image while
        allocating more computation to the parts that correspond to boundaries, and to smaller and rarer classes, as shown in
        (c.3) in \textcolor[HTML]{A52A2A}{red}. Figure best viewed on a reader with zooming capability. Full details are presented in
        \cref{sec:proposed}.}
    \label{fig:method_illustration}
\end{figure}
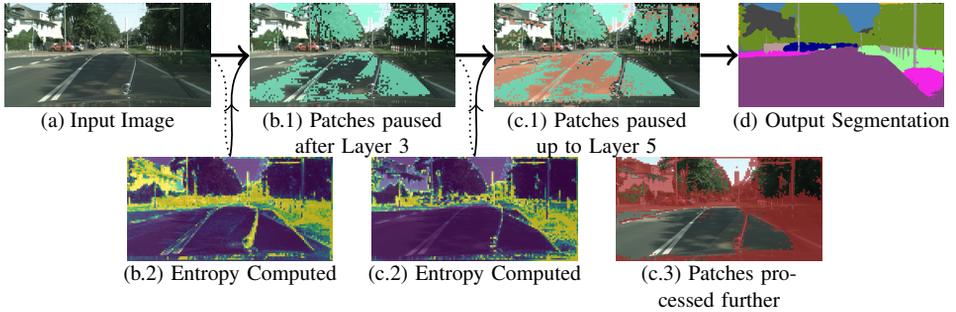

In this work, we are interested in patch dropping in the context of semantic segmentation. Differing from the case of
image classification, it is not possible to drop patches in semantic segmentation, as we have to predict the labels for
all the pixels. Instead, we redefine the problem in the context of semantic segmentation to \textit{patch pausing}: 
pausing a patch at a certain layer signifies that its representation is not going to be updated by any subsequent encoder layer,
it does not contribute to the feature computation of other patches, and it is fed directly to the decoder. Consider
segmenting natural road scenes from Cityscapes\citep{cordts2016cityscapes} in \cref{fig:method_illustration}, it is
apparent that some parts of the scene are relatively simpler to segment (say, the sky, and the road). So, we allocate
lesser computation power to these patches by pausing their feature computation, and feed them as-is to the decoder to
produce the final segmentation map. Our argument is supported by the findings in \citet{raghu2021do}; they find that the
representations of tokens is primarily modified in the first half of the network, and relies on the residual connections
to only marginally refine them in the later stages. This opens up an opportunity to reuse the representations, instead
of recomputing them, and thus improving the efficiency of segmentation transformers.

Our criterion for token pausing is the time-tested posterior entropy of the segmentation labels. We find that entropy is
strongly indicative of lower error. Our method, called Patch pAUsing segmentation transforMER (\Paumer), adds a simple
linear auxiliary decoder to predict labels and compute entropy, and processes only the patches whose class prediction is
of high entropy, \ie the network is not confident about predicting the labels of these patches, and processes them more.
Based on the {Segmenter}\citep{strudel2021segmenter} architecture, we show the performance of our method on the standard
benchmark suite of ADE20K\citep{zhou2017scene} and Cityscapes\citep{zhou2017scene}. Our method pushes the pareto front
of the speed-accuracy trade-offs, and we find that we can operate at a $50\%$ higher throughput with a drop in mean
intersection of union of $4.6\%$ and $0.65\%$ on ADE20K and Cityscapes respectively, and for doubling the throughput, the
drop is about $10.7\%$ and $4.5\%$ respectively. 

%% file: figures/example1.tex
\begin{tikzpicture}[inner sep=0pt, scale=1]
        \node (img) at (0,0){\includegraphics[width=.13\textwidth]{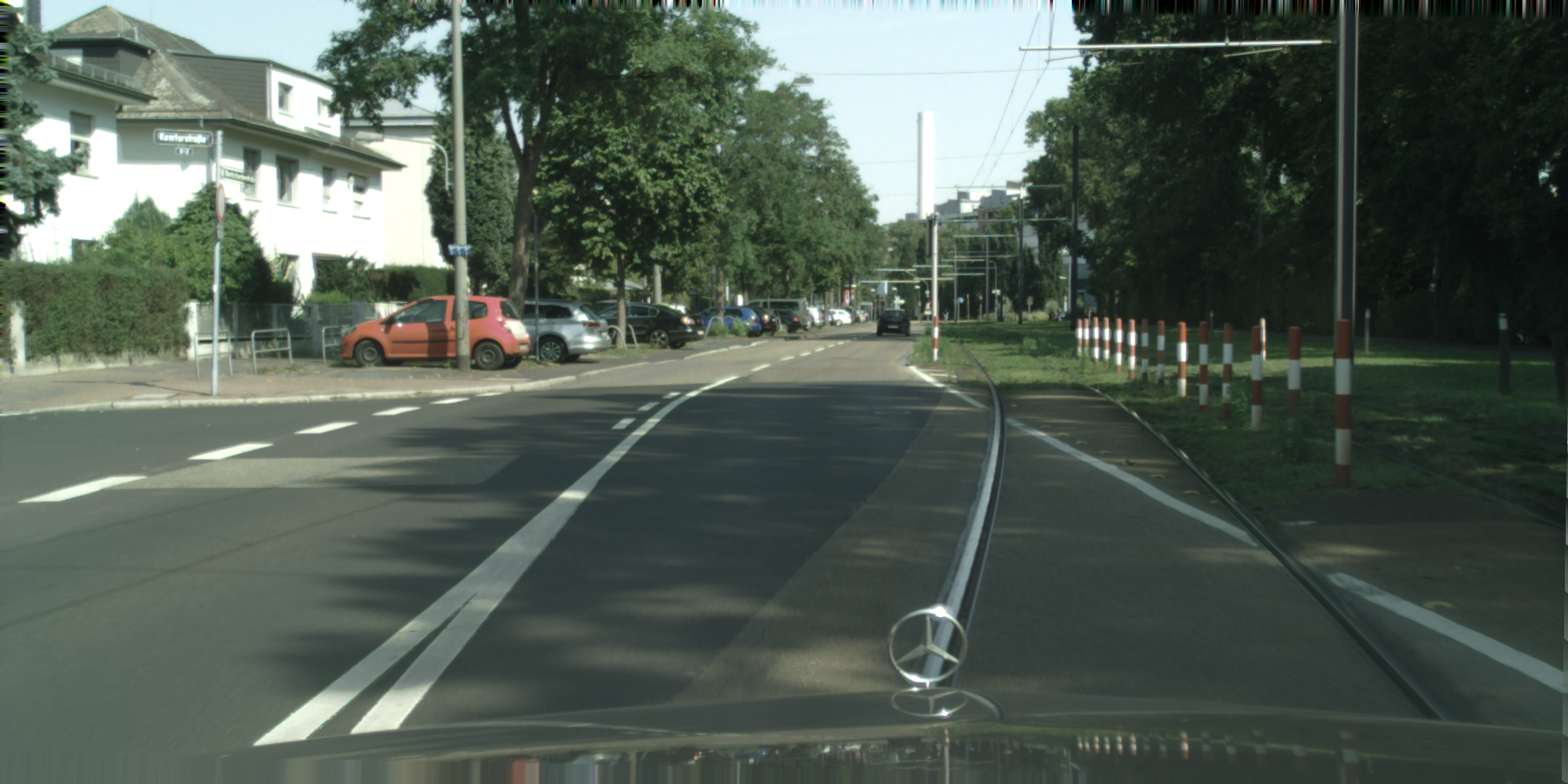}};
        \node (dr1) at (2,0){\includegraphics[width=.13\textwidth]{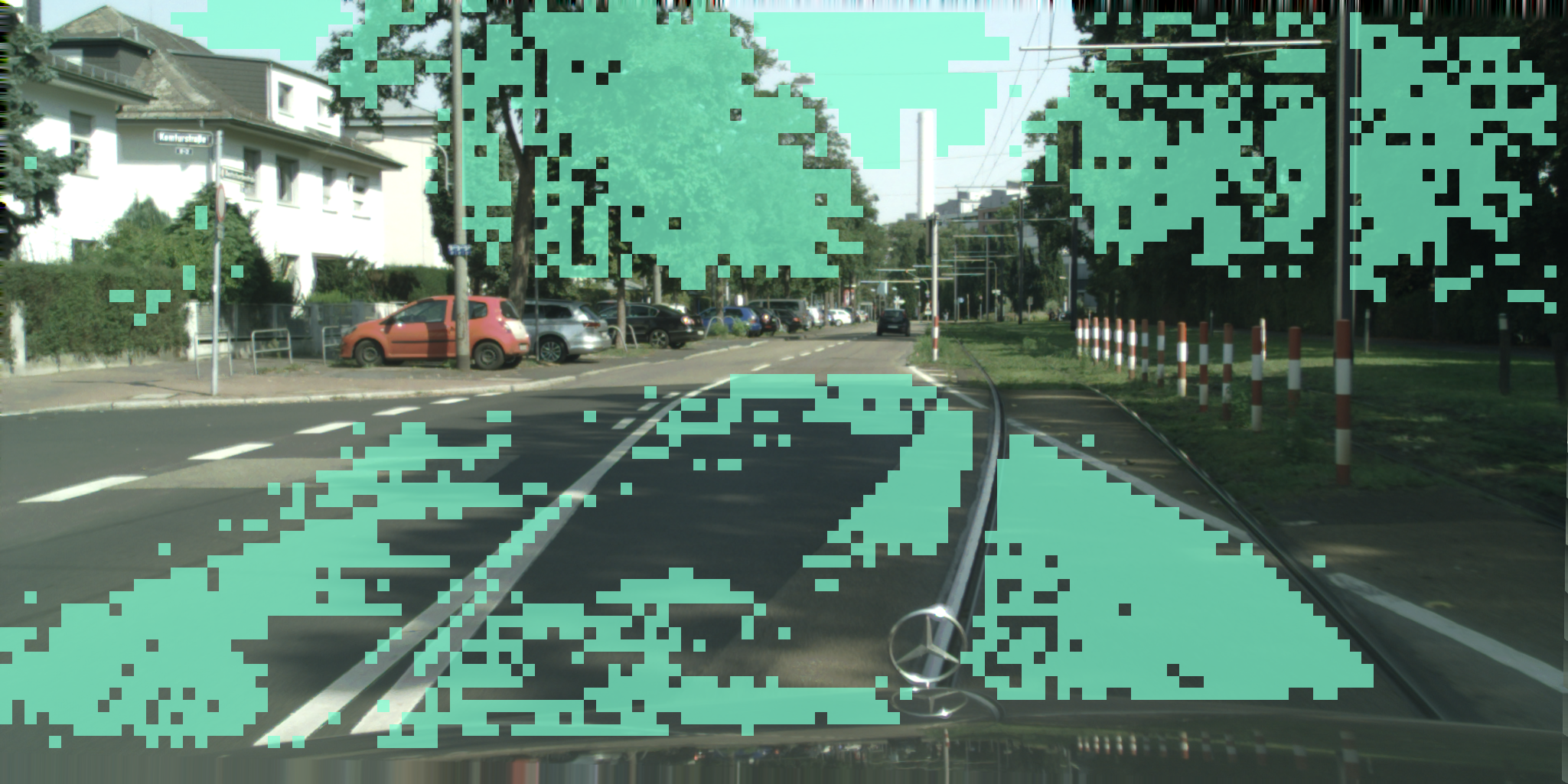}};
        \node (dr2) at (4,0){\includegraphics[width=.13\textwidth]{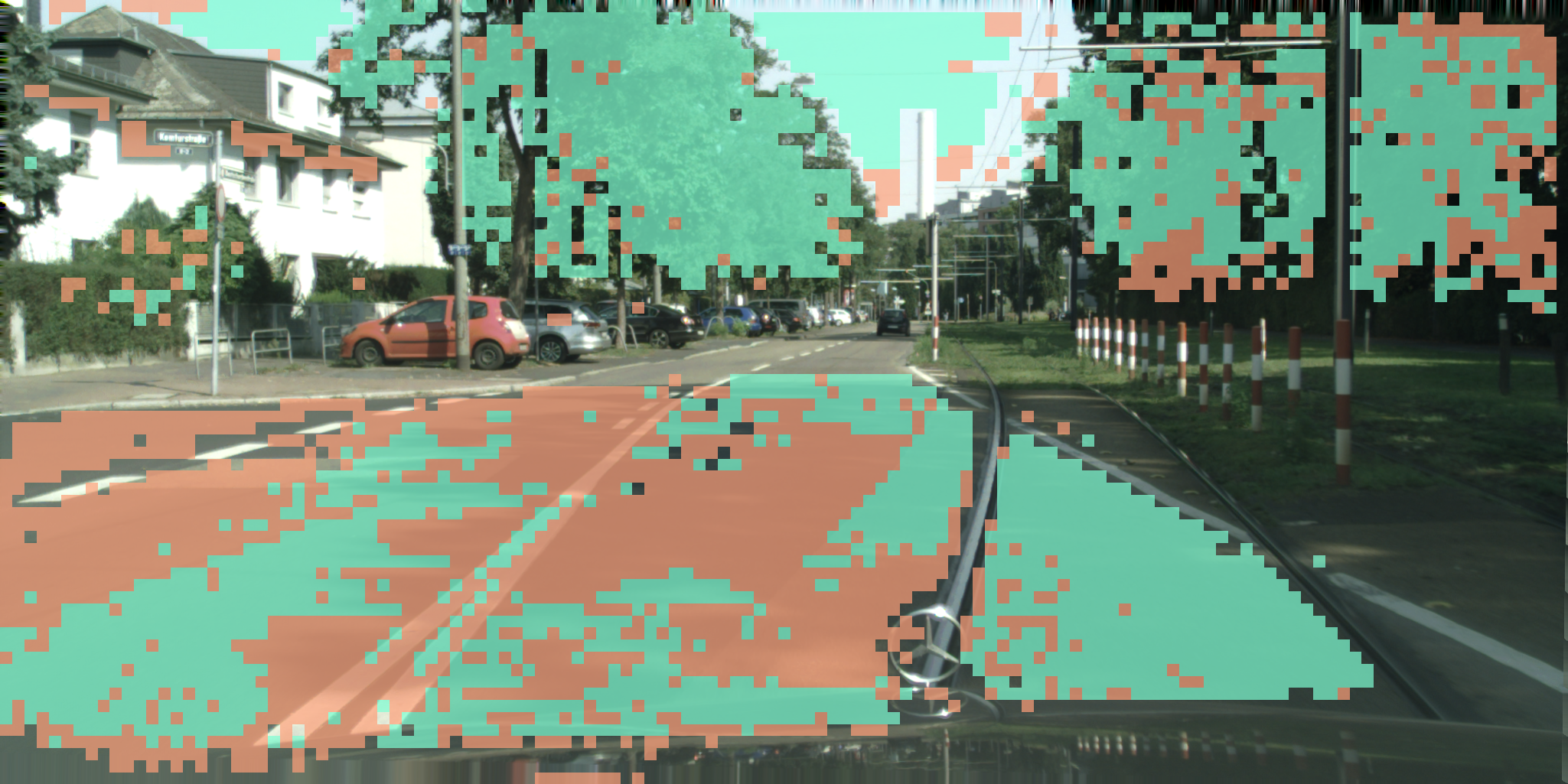}};
        \node (out) at (6,0){\includegraphics[width=.13\textwidth]{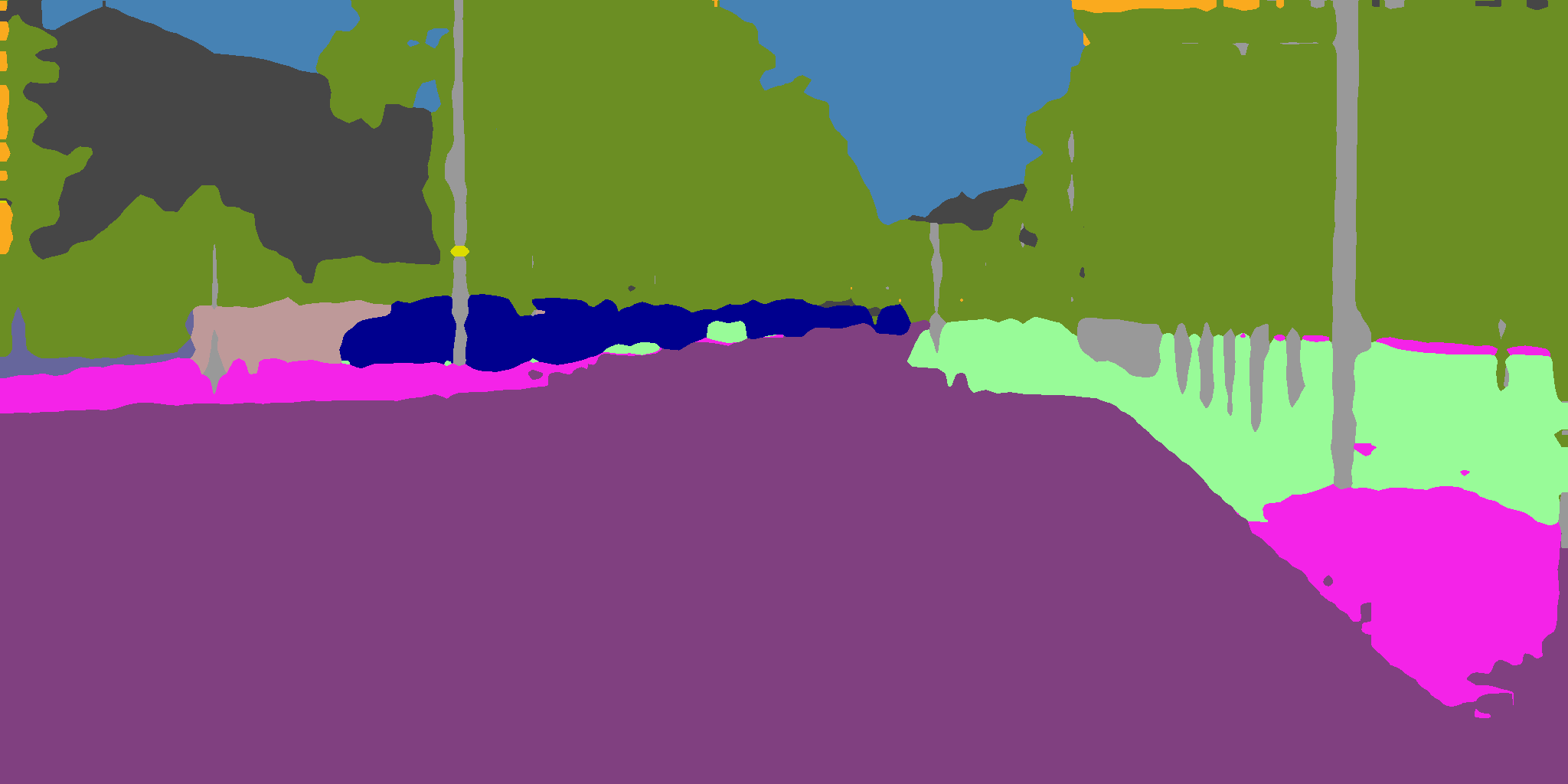}};
        \node (firstmid) at (1, 0) {};
        \node (secondmid) at (3, 0) {};
        \node (thirdmid) at (5, 0) {};
        \node[below=.8cm of firstmid.south] (ent1){\includegraphics[width=.13\textwidth]{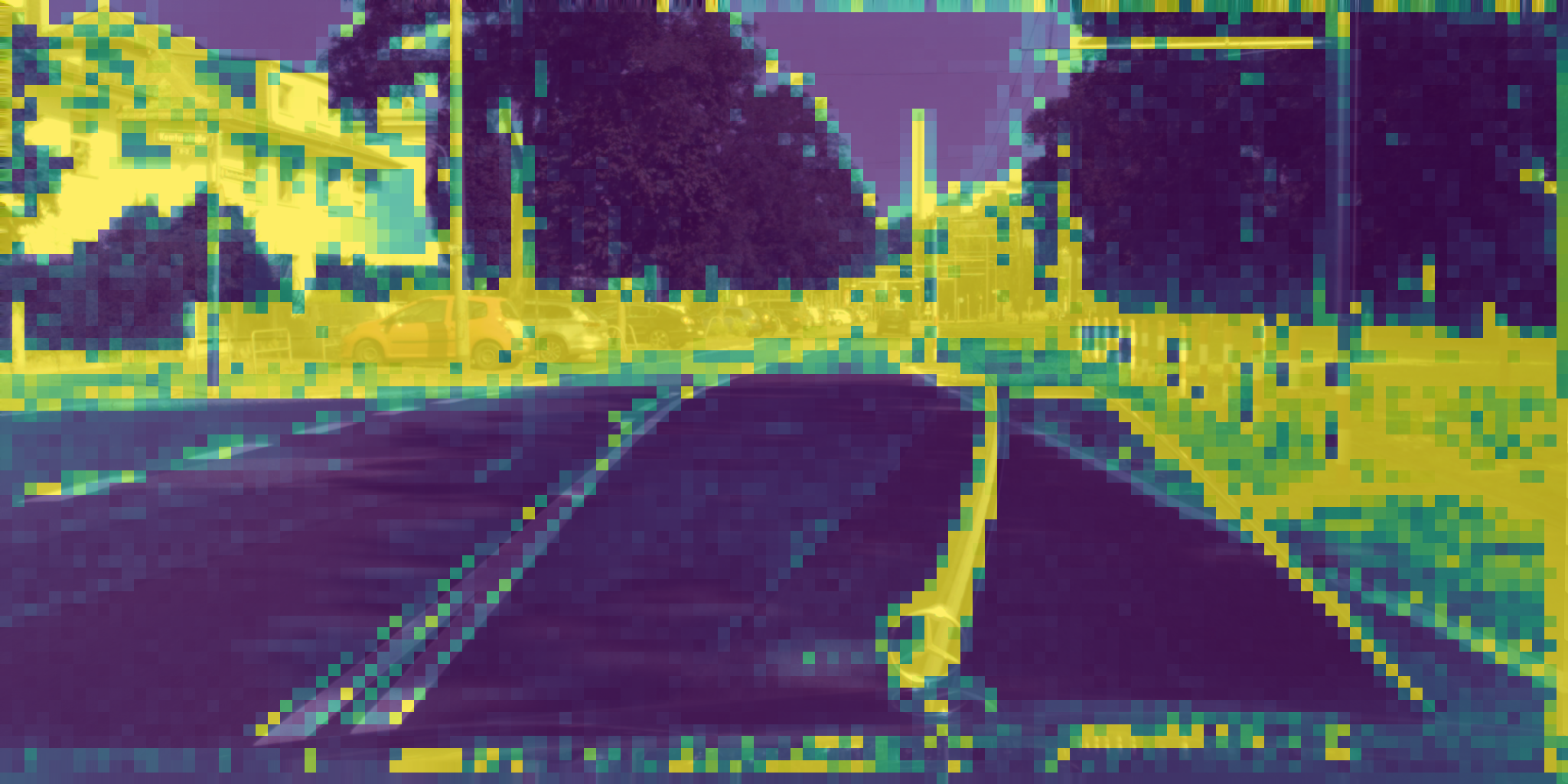}};;
        \node[below=.8cm of secondmid.south] (ent2){\includegraphics[width=.13\textwidth]{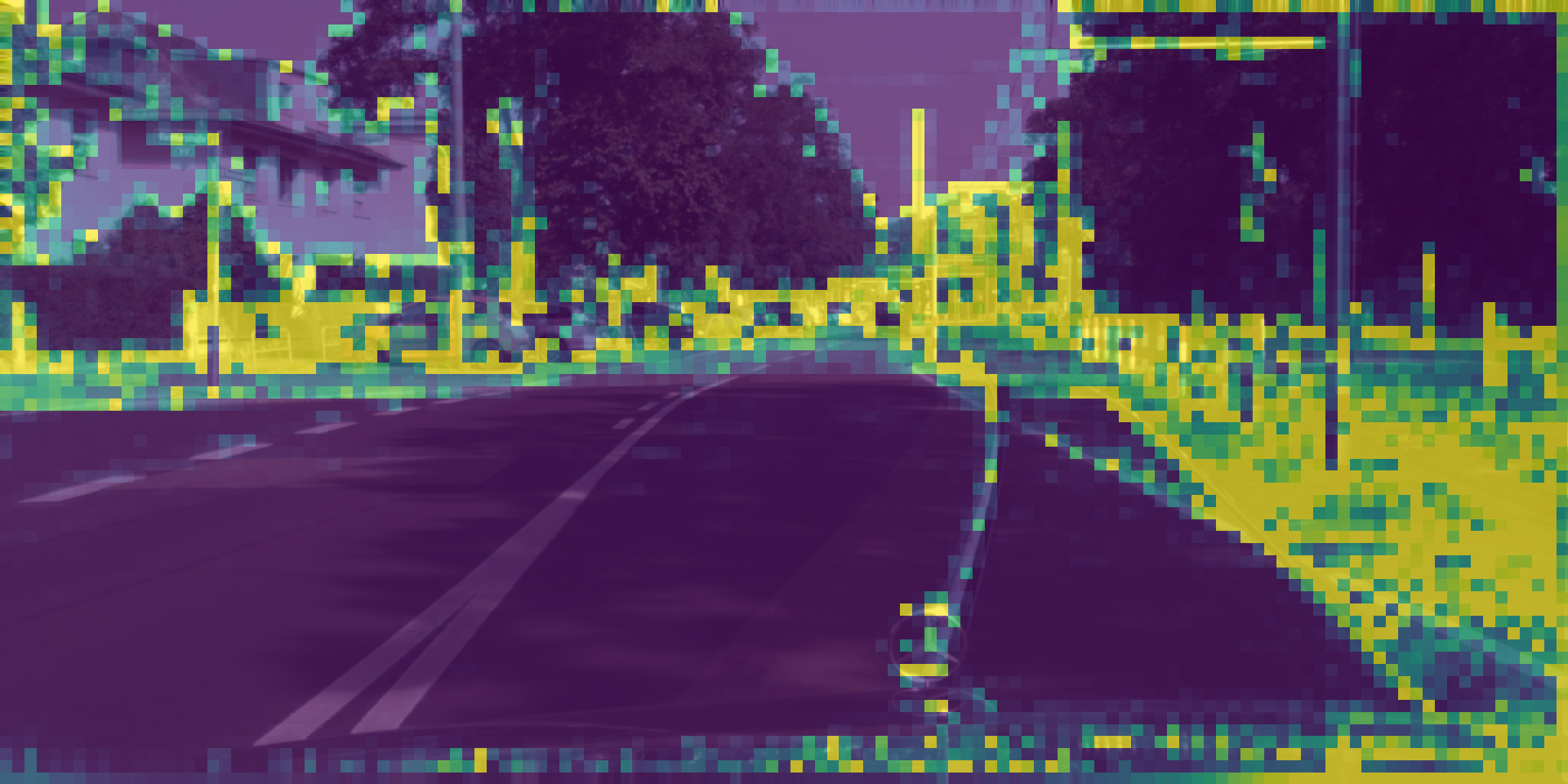}};
        \node[below=.8cm of thirdmid.south] (toproc){\includegraphics[width=.13\textwidth]{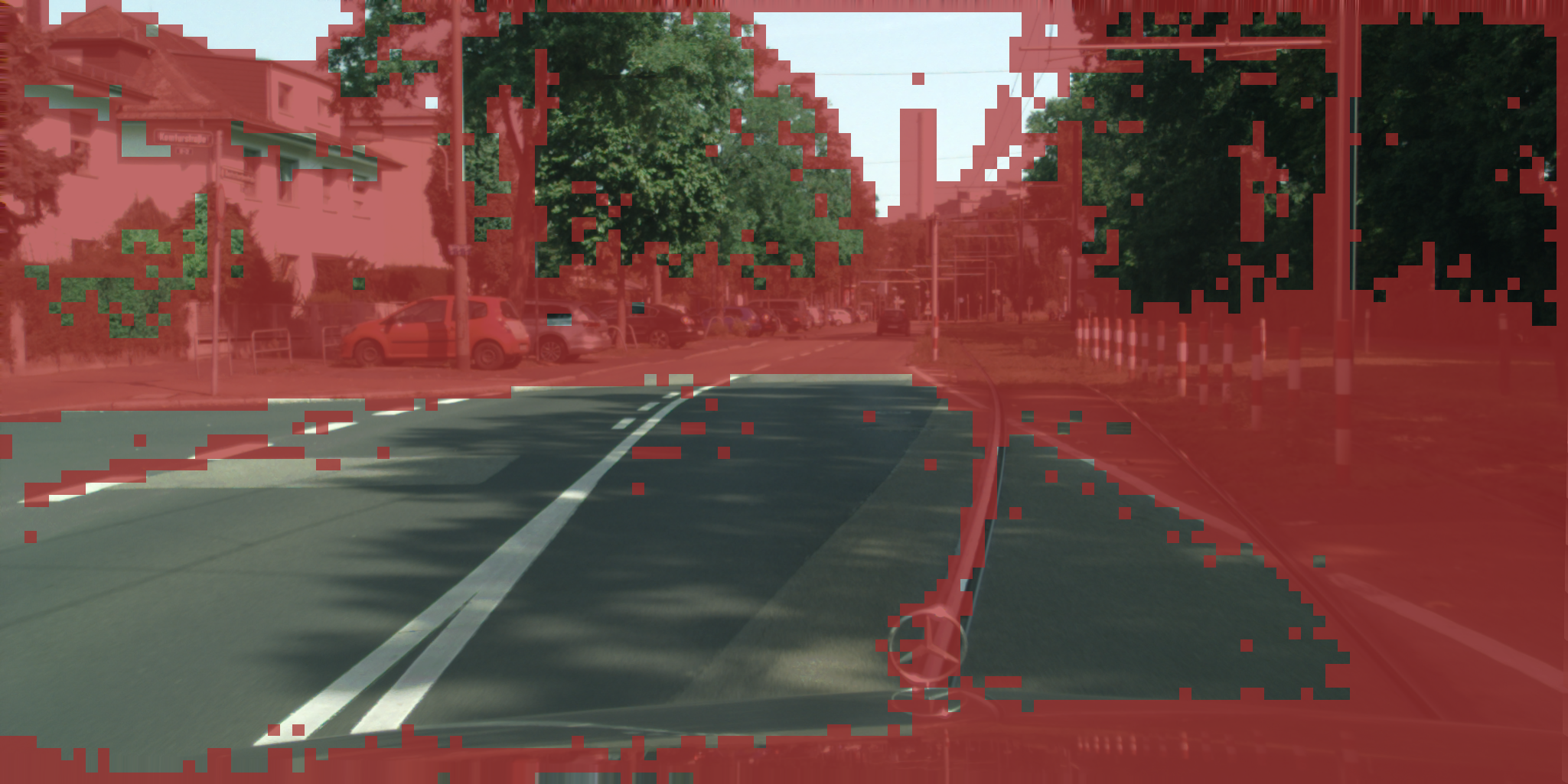}};
        \draw[->, thick] (img.east) -- (dr1.west);
        \draw[->, thick] (dr1.east) -- (dr2.west);
        \draw[->, thick] (dr2.east) -- (out.west);
        \node[below=0.03cm of img.south, font = {\tiny}, text centered] {{(a) Input Image}};
        \node[below=0.03cm of dr1.south, font = {\tiny}, text width=0.15\textwidth, text centered] {\centering{(b.1) Patches paused
                        after Layer 3}};
        \node[below=0.03cm of dr2.south, font = {\tiny}, text width=0.15\textwidth, text centered] {\centering{(c.1) Patches paused
                        up to Layer 5}};
        \node[below=0.03cm of out.south, font = {\tiny}, text width=0.15\textwidth, text centered] {\centering{(d) Output
                        Segmentation}};
        \node[below=0.03cm of ent1.south, font = {\tiny}, text width=0.15\textwidth, text centered] {\centering (b.2) Entropy
                Computed};
        \node[below=0.03cm of ent2.south, font = {\tiny}, text width=0.15\textwidth, text centered] {\centering (c.2) Entropy Computed
        };
        \node[below=0.03cm of toproc.south, font = {\tiny}, text width=0.15\textwidth, text centered] {\centering (c.3) Patches
                processed further};
                \begin{scope}[decoration={
                        markings,
                        mark=at position 0.5 with {\arrow{>}}}
                        ] 
                    
                \draw[densely dotted] (img.east) to[out=-45, in=110] (ent1.north);
                \draw[-, postaction={decorate}] (ent1.north) to[out=80, in=-135] (dr1.west);
                \draw[densely dotted] (dr1.east) to[out=-45, in=110] (ent2.north);
                \draw[-, postaction={decorate}] (ent2.north) to[out=80, in=-135] (dr2.west);
        \end{scope}
\end{tikzpicture}

%% file: proposed_method.tex
\section{Patch pausing transformer for Semantic Segmentation}\label{sec:proposed}
\begin{figure}[htpb]
    \centering
    \includegraphics[width=0.9\linewidth]{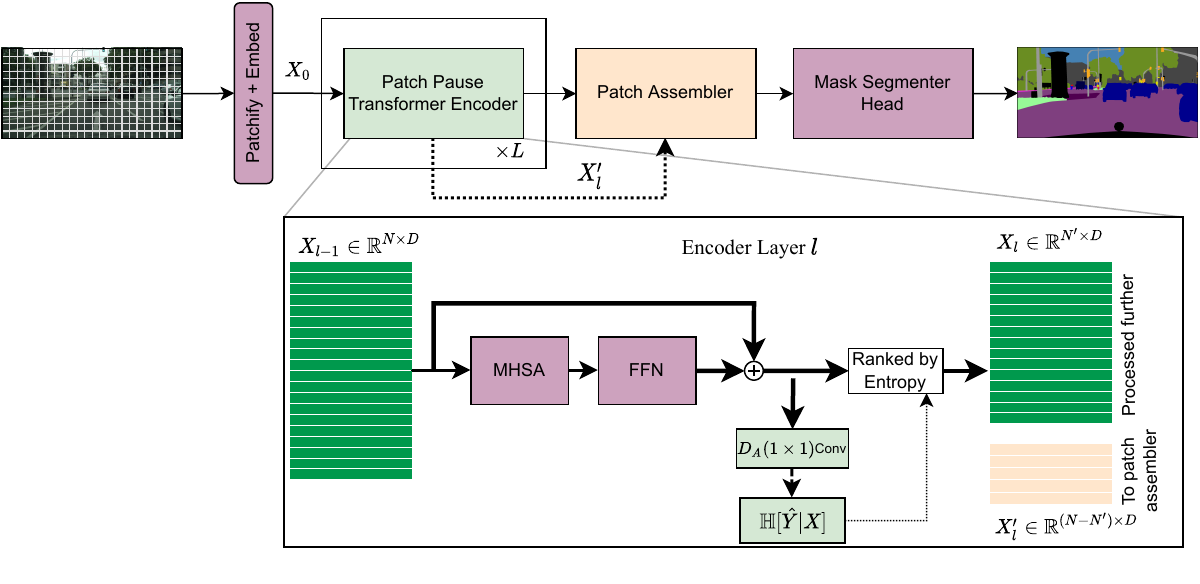}
    \caption{Schematic of our proposed method. We modify Segmenter\citep{strudel2021segmenter} to enable pausing of patches,
        and feeding them directly to the decoder. Our proposed \paumer\,encoder adds a simple auxiliary decoder (a $1\times 1$
        convolution), and uses the predicted posterior entropy ${\mathbb{H}}[\hat{Y}|X]$ of each component of $X$ to reorder the
        feature representation $X$. A portion (\pauseprop) of this feature representation would be paused and fed to the decoder
        directly. The rest of the features (of size $N' < N$) are processed further.}
    \label{fig:overall_method}
\end{figure}

Patch pausing for semantic segmentation is tied to the notion that computation should be non-uniformly distributed
across the image, with some parts of the input needing more computation than others to obtain an accurate segmentation.
This notion is difficult to realize in the case of convolutional networks as convolution implementations in popular deep
learning frameworks handle only inputs with uniform coordinate grid, and thus need software
optimizations\citep{lavin2016fast}, or require architectural simplifications like the use of $1\times 1$ convolutions in
the network\citep{verelst2020dynamic}. On the other hand, ViTs are ideally suited for this purpose, as each transformer
layer consumes a matrix of patch representations without any regard to its inputs spatial location. Removal of patches
from computation does not require any additional modifications to the transformer networks for them to apply
heterogeneously distributed computation across an image. This restricts the pausing pattern to operate at a patch level,
and these paused patches can be non-uniformly distributed over the image coordinate grid. 
Our primary experiments are based on the architecture Segmenter\citep{strudel2021segmenter}, which uses a ViT backbone
to extract patch representations, and predicts a segmentation map using a transformer-based mask decoder. We describe this in detail in
\cref{app:sec:method:segmenter}
\vspace{-10pt}

\subsection{Using Entropy as a criterion for patch-pausing}\label{sec:whyentropy}

How do we determine which tokens to pause, \ie which one do not need more processing? Consider the unrealistic case when
we have access to the ground truth labels. We could decode after each layer $l \in \{1\dots L\}$, and stop the
processing of tokens for which the prediction is accurate enough. 

\begin{figure}[htpb]
    \centering
    \includegraphics[width=\linewidth]{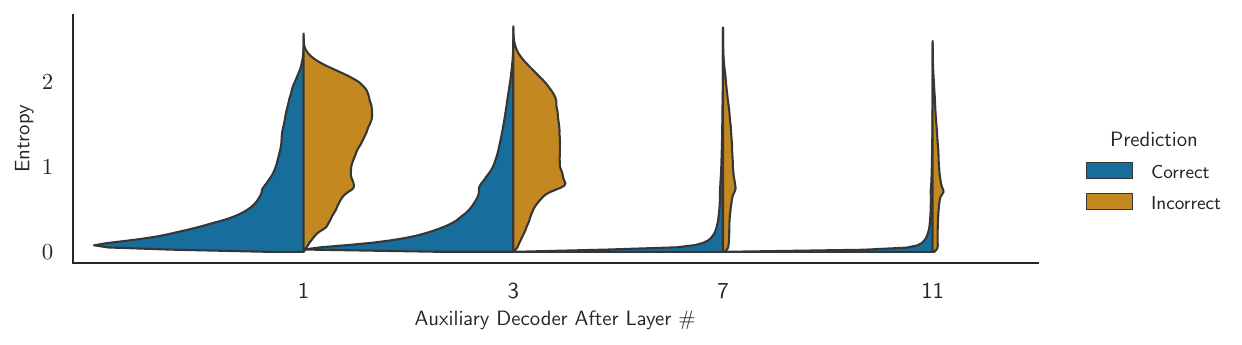}
    \caption{Violin plot of entropy of predictions at intermediate layers. In this figure, we plot the
        entropy distribution of the auxiliary predictions for $10\%$ of images in Cityscapes validation set. The x-axis marks
        after which layer the prediction was done. For each layer, the entropy distribution is shown for tokens correctly (in
        blue) and incorrectly (in orange) classified. We see that the entropies of the predictions for tokens correctly
        classified accumulate in the low values in the later layers (blue spike on the bottom-right)}
    \label{fig:ent_perf}
\end{figure}

In the absence of ground truth to determine which tokens can be paused, we propose to use the entropy of label
predictions as a proxy for the correctness of the network's predictions. 
We posit that our models, when confident about their prediction, are likely to be correct. 
To sanity-check the aptness of entropy as a pausing criterion, we plot in \cref{fig:ent_perf} the entropy of predictions
computed after every second layer in a segmenter. Specifically, we use a Segmenter with ViT-Ti backbone pretrained on
Cityscapes, freeze its weights, and only train the one-layer linear auxiliary predictor added after each layer. Each
vertical plot is a histogram, and we see that in the initial layers, there is a higher overlap of correct and incorrect
predictions' entropies. When a token has been processed enough to predict the correct label, the entropy of the
prediction is generally low. Thus, pausing patches based on entropy results in representations that have been refined
enough to result in correct predictions.

\vspace{-20pt}

\subsection{Training \paumer - One training for many pause configurations}\label{sec:trainingpaumer}

We base our network on Segmenter's architecture\citep{strudel2021segmenter} detailed in \cref{app:sec:method:segmenter}.
A pause configuration (or configuration) refers to the proportion of patches paused at each layer of the network. An
obvious method is to train and test the same patch pausing configuration that satisfies our run-time requirements. Any
changes to the run-time requirements requires retraining the network. For this reason, we propose a more general strategy that
enables multiple patch pausing configurations at inference with just one trained model. For each transformer layer $l$,
we define a range of patch pausing proportions $(\pauseprop_l^{\mathrm{lo}}, \pauseprop_l^{\mathrm{hi}})$. For each
batch of training samples, we sample uniformly one layer $l \in \{3,\dots, L\}$ and a patch pausing proportion
$\pauseprop_l\sim \mathcal{U}[\pauseprop_l^{\mathrm{lo}}, \pauseprop_l^{\mathrm{hi}}]$, where $\mathcal{U}$ refers to a
uniform distribution over the parameters.
To facilitate the patch
pausing, we employ a single auxiliary decoder $D_A$, parametrized by a $1\times 1$ convolution, after the operations of
layer $l$ (see \cref{fig:overall_method}).

The outputs of the main branch of the network and the auxiliary branch are trained using the traditional cross entropy
loss.
\begin{align}\label{eqn:main_loss}
    \loss_{\text{main}}  & = \text{CE}(\bm{y}, \bm{\hat{y}})                       \\
    \loss_{\text{aux}}^l & = \text{CE}(\bm{y}, \bm{\hat{y}^l})\label{eqn:aux_loss}
\end{align}
where $\bm{y}$ is the ground truth, $\bm{\hat{y}}$ refers to the logits predicted by the main decoder (mask transformer),
and $\bm{\hat{y}^l}$ is the auxiliary decoder's output at the l\textsuperscript{th} layer.
The total loss used to train is a combination of losses in \cref{eqn:main_loss,eqn:aux_loss}.
\begin{equation}\label{eqn:total_loss}
    \loss_{\text{total}} = \loss_{\text{main}} + \lambda  \loss_{\text{aux}}^i
\end{equation}
where $\lambda$ is a scalar used to scale the contribution of additional losses.

At layer $l$, entropy is computed for each component of $X_l$ as
\begin{equation}\label{eqn:ent_predict}
    H_l \coloneqq \mathbb{H}[\hat{Y}_l | X_l] = \mathbb{H}[\sigma(D_A(X_l))],
\end{equation}
where $\sigma$ is the softmax function applied to each pixel independently. With this entropy, we pause the computation
of a proportion $\pauseprop_l$ of tokens with the lowest entropy and store them as $X'_{l}$, and continue with the computation using
the rest of the tokens $X_{l}$ (see \cref{fig:method_illustration}). Note that there are other ways to use $H_l$ to
pause tokens, for example by using a threshold on entropy. However, doing so results in pausing and removing different
amounts of tokens in each image of a batch, which would add a substantial overhead as padding is not processed efficiently on a GPU. Additionally, pausing a fixed amount of tokens allows for a deterministic computation time.

In order to re-assemble in the original order the patches that have been fully processed and the ones that have been
paused, we use a patch assembler module (see \cref{app:sec:assemblerps} for PyTorch-like code). It takes $X_L$ and
$X'_l$, and reassembles them into the original shape of $X_0$. In order to do so, pausing mechanism stores the indices
of the patches that have been chosen to be paused, in addition to the current feature representation. The assembler
copies the paused representation into the same indices stored previously. This re-assembled output is finally fed into
the decoder (mask segmenter head) to compute the segmentation map.

\textbf{Inference}
Our training procedure of randomized pause configurations gives us the advantage
to choose a pausing configuration that is informed by the run-time requirements \ie mIoU and number of images processed per
second. This configuration can have multiple pause locations, each with different pause proportions. We show some results for
some configurations listed in \cref{table:exp_configs}. The patch assembler accordingly assembles multiple paused
patches $\{X'_l\}$, and the final representation $X_L$. The specific configurations are chosen to display the
adaptability of the trained network to various inference time requirements, and do not hold any specific importance.
Pause configurations can be added easily, as it does not influence the training, but only requires testing over the
validation set. 


%% file: previous_work.tex
\section{Related Work}\label{sec:prevwork}
We now discuss some important prior work related to our method, before showing the experimental evidence.

\textbf{Segmentation using Transformers}
Transformers that were originally proposed for language processing tasks\citep{vaswani2017attention} have been
incorporated into vision\citep{dosovitskiy2020vit,touvron2021deit} and several improvements have been
proposed\citep{salman2021transformerssurvey}. SETR\citep{zheng2021rethinking} adapted the standard vision transformer
(ViT) to segmentation by using multiscale decoder on all the image patches. Segmenter\citep{strudel2021segmenter}
improved the decoder design by using a learnable per-class token that acts as weighting mechanism over the tokens'
representations. Segformer\citep{xie2021segformer} redesigned the architecture with a multiscale backbone that does not
use positional encoding, and an MLP based decoder. Several improvements to the transformer backbone have been shown to
have impact on the down-stream segmentation performance\citep{wang2021pyramid,xu2021co,chu2021twins,liu2021swin}. These
improvements to the transformer backbones have indeed improved the efficiency, measured by frames processed per second,
number of floating point operations per second (FLOPs), or images processed per second. 

\textbf{Token sparsification methods}
Several components of the whole transformer architecture have been improved, by approximations, and simplifications to
attention mechanism. Interested readers can refer to \citet{tay2020efficient}. Orthogonal to the architectural
improvements, recent work has focused on the reduction of the data processed, and our proposed method is a form of input
dependent reduction. \citet{graves2016adaptive} proposed Adaptive Computation Time (ACT), where the amount of processing
for each input to an RNN is decided by the network by determining a halting distribution. It was adapted to residual
networks by \citet{figurnov2017spatially}, that dynamically decides to apply differential number of residual units to
different parts of the input. This has been adapted to transformers too\citep{yin2022avit}, where the tokens are
progressively halted as they are determined have been processed enough according to a similar criterion as ACT.
DynConv\citep{verelst2020dynamic} uses an auxiliary network to predict pixel masks using which indicate pixels of the
image that are not processed by a residual block. DynamicViT\citep{rao2021dynamicvit} extends this formulation to
transformers where they, similarly, use an auxiliary network to predict which patches are dropped from being refined
further. The auxiliary branches are trained using the Gumbel-softmax trick\citep{jang2016categorical} in both these
methods. We consider simplicity of the steps the strength of our proposed method. Unlike
DynamicViT\citep{rao2021dynamicvit}, we do not need techniques like Gumbel-softmax that are harder to optimize, and
additional tailored losses. Additionally, both A-ViT and \dvit drop a fixed amount of patches for a given image, and do
not provide the flexibility to vary the number of patches dropped, as our method does.

\textbf{Early-exit methods} 
Dynamic neural networks\citep{han2021dynamicnn} adapt the architectures or parameters in an input adaptive fashion.
Specifically, early-exit methods find that deep neural networks can {overthink} where a network can correctly predict
before all layers process the input, and it can even result in wrong predictions due to
over-processing\citep{kaya2019shallow}. Several methods to determine when to exit the network have been proposed.
Branchynet\citep{teerapittayanon2016BranchyNet} and Shallow-Deep nets \citep{kaya2019shallow} uses auxiliary classifiers
to predict the output class for vision convolutional networks, and stops processing a sample if the entropy of a
branch's predictions is lower than a predefined threshold. This idea was further exploited in NLP literature.
\citet{zhou2020bert} extends this by using a patience parameter that tracks number of auxiliary classifiers which
predict the same class. DeeBERT\citep{xin2020Deebert} proposes a two stage training, where the auxiliary decoders are
trained after the main networks is trained, and frozen. \citet{li2017not} propose a layer cascade for convolutional
segmentation networks, that processes easy to hard parts progressively through the network. Their method needs
modifications of the network architecture, whereas we show that our proposed method can be added with very little
efforts. Our method is an early exit strategy, specifically for the case of segmentation transformers. While similar
methods have been examined in literature, to the best of our knowledge, we are the first to examine the patch pausing
problem of semantic segmentation. Also, the randomized training presented in \cref{sec:trainingpaumer} has not been used
in this context, though similar ideas to reduce network width were studied in slimmable networks\citep{yu2018slimmable}.



%% file: experiments.tex
\section{Experiments}\label{sec:experiments}
\subsection{Datasets and Evaluation}\label{sec:exps:subsec:datasets}

We show the performance of our method using networks trained on Cityscapes\citep{cordts2016cityscapes} and
ADE20K\citep{zhou2017scene}. Cityscapes (CS) consists of $2,975$ images in the training set, in which each pixel belongs
to one of $19$ classes, and $500$ images in the validation set which are used to benchmark the performance of our
method. ADE20K is substantially larger, with a training set of $25,574$ with $150$ classes, and $2,000$ images to
validate the performance. The results for Cityscapes and ADE20K are presented below.

Our primary performance measure is based on the  speed-accuracy trade-off, measured by mean Intersection over Union
(mIoU) metric and throughput in images per second. To determine the number of images processed per second (IMPS),
following \citet{strudel2021segmenter}, we use images of size $512 \times 512$ with a batch size that optimally occupies
a V100 GPU. 

\textbf{Methods compared}
To assess the performance of our proposed method, we use the following baselines for comparison:
\begin{enumerate}[\hspace{0.5cm}a.]
    \setlength{\itemsep}{0pt}
    \item \textbf{Baseline set by Segmenter}, without any patch pausing.\vspace{-1pt}
    \item \textbf{Random Pausing (RP)}: We train the network to handle pausing a proportion \pauseprop\xspace of
          randomly chosen patches, instead of the lowest entropy ones.\vspace{-1pt}
\end{enumerate}
We examined an additional simple baseline of random pausing (without training), and found the results not competitive
enough to warrant reporting here. Also, some methods in \cref{sec:prevwork} are capable of dropping different patches
per image. These methods can result in a decrease in FLOPs (floating point operations), but this reduction cannot be
realized in wall clock improvements as modern GPUs parallelize computation over batch elements.

\newcolumntype{g}{>{\columncolor{gray!25}}l}

\begin{table}[htpb]
    \centering
    \begin{tabular}{@{}c|l|lglglglglglgl@{}}
        \toprule
        \multirow{4}{*}{\rotatebox{90}{Pause Layer}} &   & \multicolumn{13}{c}{Pause configurations}                                                                         \\ \cmidrule(l){2-15}
                                                     & 3 & 0.2                                       & 0.4 & 0.6 &     &     &     &     & 0.2 & 0.3 & 0.4 & 0.2 & 0.3 & 0.4 \\
                                                     & 5 &                                           &     &     & 0.2 & 0.4 & 0.6 & 0.8 & 0.2 & 0.3 & 0.4 & 0.2 & 0.3 & 0.4 \\
                                                     & 7 &                                           &     &     &     &     &     &     &     &     &     & 0.2 & 0.3 & 0.4 \\ \bottomrule
    \end{tabular}
    \vspace{10pt}
    \caption{Table of configurations. Each column represents a pause configuration, \eg the first column represents the
        configuration 
        pausing $20\%$ of tokens after layer 3, using the notation introduced in
        \cref{sec:trainingpaumer}. Each configuration here corresponds to a marker in \cref{fig:mious_tradeoff_plot,fig:dropincomparison,fig:ade_miou_tradeoff}.}
    \label{table:exp_configs}
\end{table}

\textbf{Training hyperparameters and Inference Configurations:}
For our main experiments, we use Segmenter with ViT-Ti and ViT-S backbones (details in \cref{app:sec:backbonedeets}).
During training, we follow the procedure in \cref{sec:trainingpaumer} for every network and dataset, and in particular
we pause a random amount of tokens $\pauseprop_l \sim \mathcal{U}[0.2,0.8]$ at a random layer $l \in \{3,4,5,6,7,8,9\}$.
We initialize the model for our training with pretrained segmenter weights, as we find this results in better
performance, and train the model for $80,000$ steps for Cityscapes and $160,000$ steps for ADE20K. The auxiliary
loss-weight $\lambda$ is set to $0.1$. Rest of the hyperparameters as kept the same as in \citet{strudel2021segmenter}.
We implement our method using mmsegmentation\citep{mmseg2020}, and use their pretrained models whenever available. At
inference, we test the networks with the pause configurations in \cref{table:exp_configs}. This list of pause
configuration is not exhaustive, and does not hold any specific importance, but have been chosen to show the efficiency
of our method in trading off mIoU for higher IMPS.

\textbf{Choosing pause configurations:}
Determining the appropriateness of a pausing configuration incurs little additional cost, as it only requires inference
with a validation set for each configuration of interest (see \cref{fig:ade_miou_tradeoff,fig:mious_tradeoff_plot}). On
a new dataset for which we train the network with the proposed training procedure, we foresee two scenarios:
\begin{enumerate}
    \item If the objective is to attain a specific throughput, we can easily find configurations that match the requirement with a sweep
over them (\cref{app:fig:droponce,app:fig:paretofronttwice}) by only timing them, and then evaluating the mIoU of the
ones that meet the time requirement.
\item If the objective is to find a good throughput-mIoU trade-off: First, we sweep
through configurations that pause at only one layer (\cref{app:fig:droponce}) and we pick the first layer and
proportion. We fix this layer and ratio, then sweep through pausing configurations at a second layer
(\cref{app:fig:paretofronttwice}). We repeat this procedure until adding more layers is no longer beneficial.
\end{enumerate}



\subsection{Results}

\paragraph{Performance analysis}\label{sec:exps:subsec:perf}

\begin{figure}[htpb]
    \centering
    \begin{subfigure}{\linewidth}
        \includegraphics[width=.95\linewidth, keepaspectratio]{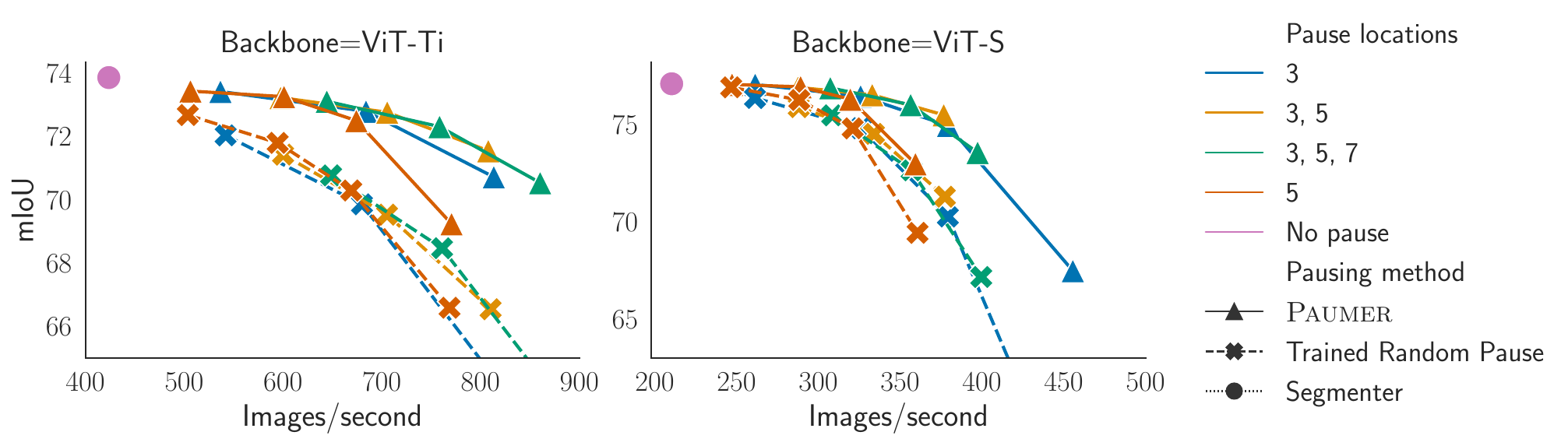}
        \caption{mIoU vs Images processed per second for ViT-Ti and ViT-S backbones on Cityscapes val set.}
        \label{fig:mious_tradeoff_plot}
    \end{subfigure}
    \begin{subfigure}{\linewidth}
        \includegraphics[width=.95\linewidth]{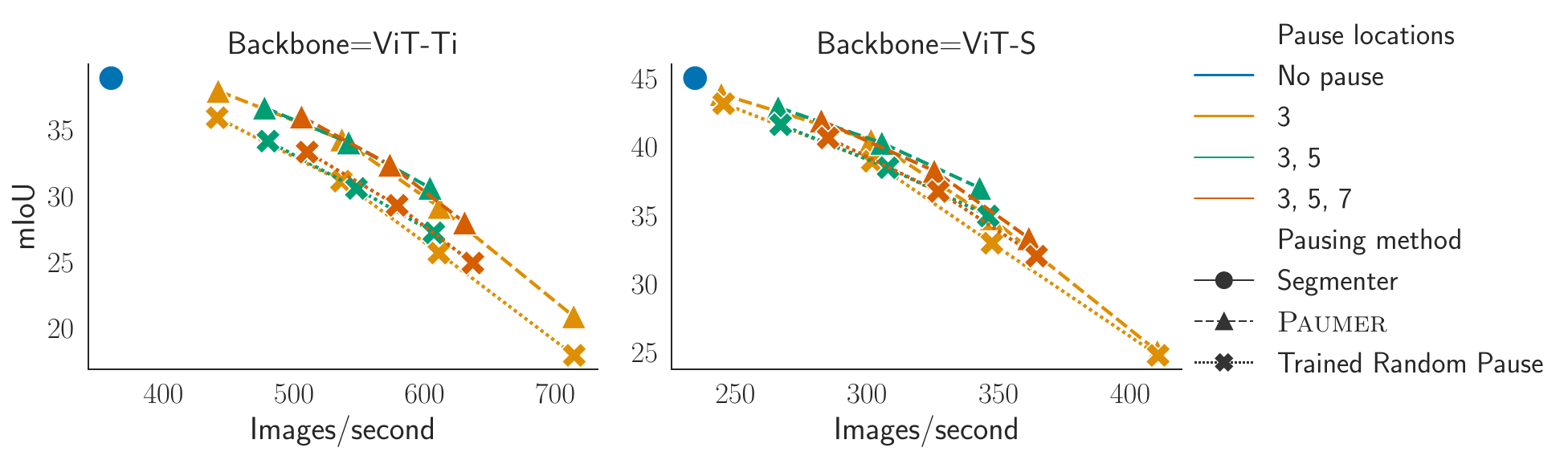}
        \caption{mIoU vs Images per second for ViT-Ti and ViT-S backbones on ADE20K val set. }
        \label{fig:ade_miou_tradeoff}
    \end{subfigure}
    \vspace{2ex}
    \caption{Results on Cityscapes and ADE20K for our proposed method \Paumer.  Each marker is a configuration from \cref{table:exp_configs}. We train a single model capable of handling various pause configurations that can be
            chosen based on run-time requirements. It is apparent that ADE20K suffers from a higher drop
            in performance when patch pausing is employed. However, \paumer\, consistently outperforms the random training
            baseline.}
\end{figure}

In \cref{fig:mious_tradeoff_plot}, we plot the mIoU versus the number of images per second achieved by baselines and our
entropy patch pausing for different configurations, for Cityscapes. Each point is the average value of three training
runs. The left-most point corresponds to the original Segmenter model, in solid line our proposed pausing strategy, and
in dashed line random pausing with training. For both ViT-Ti and ViT-S backbones, a $50\%$ increase of IMPS can
be achieved with an mIoU drop of about $0.7\%$ and $0.6\%$ respectively. Further, for doubling the IMPS, we see that the
mIoU drops about $3.2\%$ and $5.9\%$ respectively. For the trained random pausing using ViT-Ti, a strong baseline, the
equivalent drops in mIoU are about $2.9\%$ and $8.8\%$ to increase the IMPS by $50\%$ and $100\%$, respectively.
We show results for ViT-Ti and ViT-S for ADE20K in \cref{fig:ade_miou_tradeoff}. For ViT-Ti backbone, we see that for a
$50\%$ increase in IMPS, mIoU drops by about $2.8\%$, and a $100\%$ increase in IMPS with a mIoU drop of about $10.7\%$,
compared to random pausing performance of $5.4\%$ and $13.5\%$. The drop in mIoU is in contrast with the results of
Cityscapes, where we could achieve similar increase in throughput with a much lesser drop in performance. We chalk this
difference up to dataset characteristics; ADE20K has almost an order of magnitude more classes, and the images are
smaller with cluttered scenes and numerous small objects, which may require more processing to be correctly classified.

Generating these performance plots \ie mIoU vs IMPS is inexpensive, as no retraining is involved and only needs
inference on a validation set with reasonably chosen pause configurations.




\paragraph{Comparison to other architectures}
\begin{figure}[htpb]
    \centering
    \begin{minipage}[b]{0.49\textwidth}
        \vspace{-10pt}
        \includegraphics[width=\textwidth]{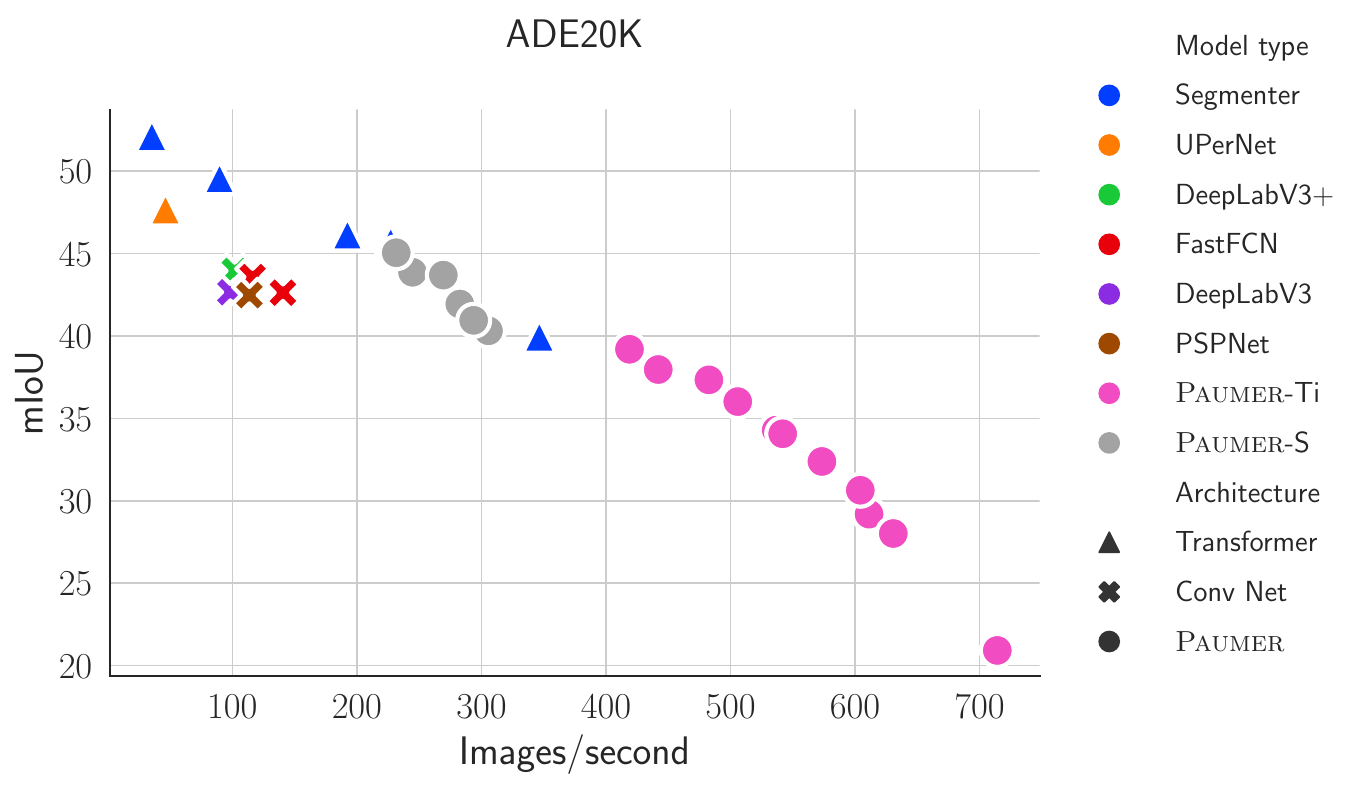}
        \label{fig:first}
    \end{minipage}
    \hfill
    \begin{minipage}[b]{0.49\textwidth}
        \vspace{-10pt}
        \includegraphics[width=\textwidth]{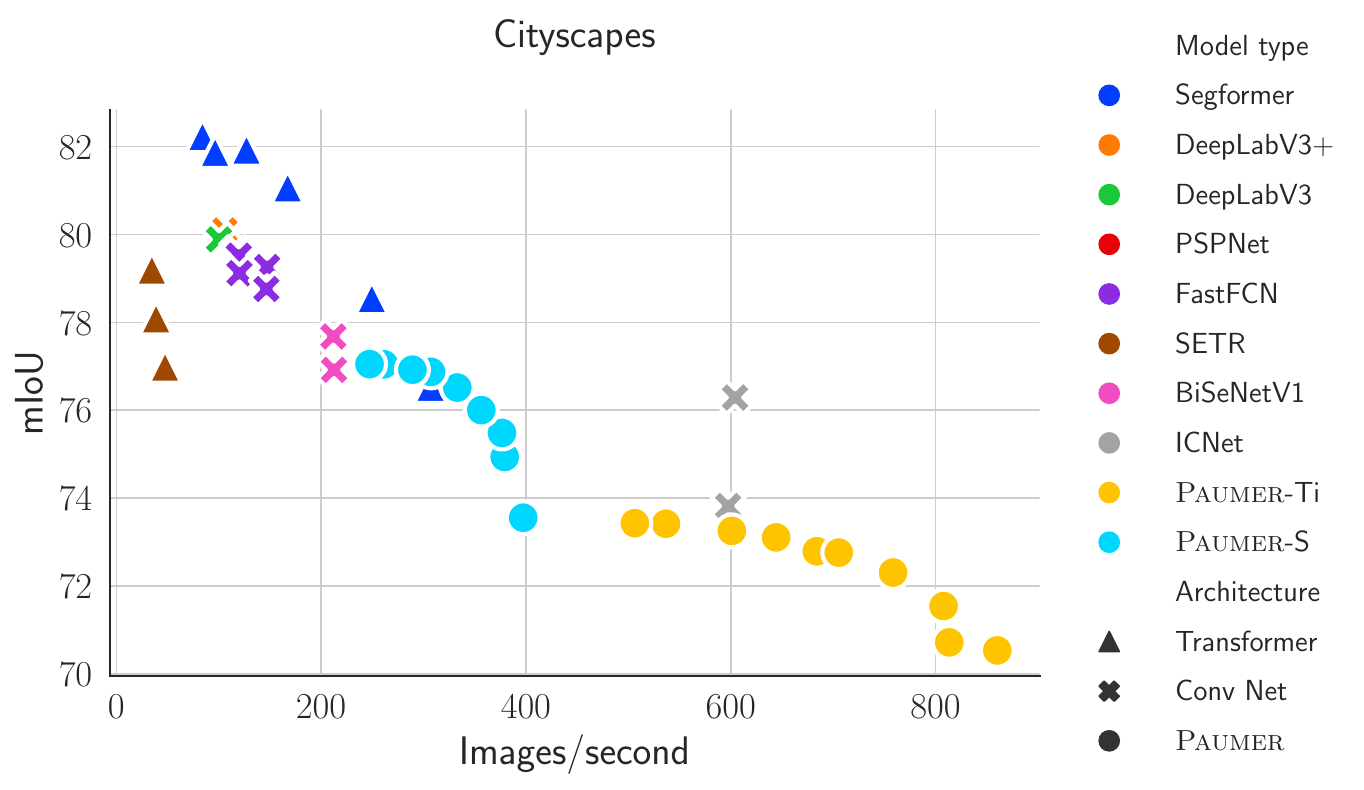}
        \label{fig:second}
    \end{minipage}
    \caption{Performance comparison for ADE20K and Cityscapes. We compare to pretrained models available in
mmsegmentation\citep{mmseg2020} for each of these datasets. Architectures devised for speed or accuracy outperform 
us on those criteria, but \paumer\xspace has the unique advantage that we can trade off one for the other using a tunable
hyperparameter.
Here, we show the pareto front of \paumer\. 
We use different colors for different architectures, and different shapes for architectural families.
    }
    \label{fig:broader_comparison}
\end{figure}
In \cref{fig:broader_comparison}, we compare our method to a broad array of architectures for which pretrained models
are available in mmsegmentation\citep{mmseg2020}.
Here we plot only the best performances obtained for each throughput across pause configurations. We estimate this by computing the skyline queries.
We include both convolution based architectures, and transformer based ones.
In the transformer family, available networks have focussed on improving the performance, and thus are slower, but
more accurate than the \paumer\, family of models. CNN based ones (say ICNet for Cityscapes) that have been designed to be more efficient are
competitive or better in speed than our models. However, we have the unique ability to tune the mIoU-throughput scores
of our model without having to retrain them. 


%% file: discussion.tex
\section{Discussion}
\label{sec:discussion}
The improvement in images processed per second is obtained by reducing the number of patches processed.
This might not necessarily hold true in the case of networks with convolutional layers interspersed, such as
SegFormer\citep{xie2021segformer} that uses convolution instead of positional encoding, as convolution on unstructured
sparse inputs is not highly optimized in CUDA implementations. Thus, our method is not readily applicable to all
transformer models. 

In addition to architecture dependence, patch pausing's performance maybe dependent on the dataset itself. We attributed
the difference between Cityscapes and ADE20K to inherent dataset complexities by examining the performance of the
auxiliary classifier; for ViT-Ti, it reaches around 60\% accuracy for ADE20K and 90\% accuracy for Cityscapes. 
Examining the possible relationship between patch pausing performance and the dataset
difficulty\citep{ethayarajh2022understanding} might shed some light on this issue.

Additionally, patch pausing assumes that a paused token is not important to the feature computation of other tokens, as
it will not contribute further to the attention computation to refine the representation of other patches. Performance
(mIoU) indicates that it might have little bearing, but this assumption needs to be investigated further. 

Our method, while being input adaptive in choosing the patches to pause, chooses a fixed proportion of them. This design is
to exploit the batch level parallelism on GPUs. Choosing the number of patches depending on the input batch has not been dealt
with in this paper.

\section{Conclusion}
Our method, \Paumer, is a first step in the direction of post-hoc design for efficient inference in semantic
segmentation transformers.
We do so by applying dissimilar amounts of computation to various patches of an input image. Patches with high predicted
auxiliary entropy are processed further, whereas the rest of them are fed directly to the decoder skipping all the
intermediate computation. To run at a specified throughput (images per second), our method offers the flexibility to
choose an appropriate pause configuration, without having to retrain the network. 


%% file: appendix.tex
\section{Pseudocode for Patch Pauser and Assembler}\label{app:sec:assemblerps}

In \cref{alg:method}, we present pseudo-code for both patch pausing mechanism and patch assembler. The code isn't meant
to be functional but only for illustrative purposes. Comments describe the functionality.
\vspace{-10pt}
\begin{algorithm}
  \caption{Patch Pauser and Assembler Pseudocode}
  \label{alg:method}
  \begin{minted}[breaklines, breakanywhere, fontsize=\footnotesize, autogobble, escapeinside=||, mathescape=true]{python}
      def patch_pauser(tokens, pause_ratio, keep_indices, paused_tokens):
          
          # tokens ($X_l$) refers to current set of tokens being processed. 
          # Note that this might not be the total number of tokens, as one
          # or more patch pausing stages could have happened.
          
          # pause ratio is $\tau$, pausing proportion

          # keep_indices, paused_tokens are temporary arrays to store 
          # details for assembling (see below).
          
          _, total_tokens, _ = tokens.shape
          to_process_count = N - int(pause_ratio * N)
          
          # Compute aux entropy of tokens
          aux_prediction = auxiliary_classifier(tokens)
          probs = aux_prediction.softmax(dim=-1)
          entropy = compute_entropy(probs)

          ## Instead of sorting, we use topk. This is faster on GPU.
          topk_inds = entropy.topk(to_process_count)
          kept_tokens = tokens[:, topk_inds]

          keep_indices.append(topk_inds)
          paused_tokens.append(tokens) ## This is $X'_l$

          return kept_tokens ## This is $X_{l+1}$
  \end{minted}
  \begin{minted}[breaklines, breakanywhere, fontsize=\footnotesize, autogobble, escapeinside=||, mathescape=true]{python} 
      def patch_assembler(X_L, paused_tokens, keep_indices):
          # X_L ($X_L$) refers to the final feature representation at the end of the
          # encoder. One or more stages of pausing have occurred before this.
          # X_L is of the shape BxN'xD.

          # paused_tokens: the feature representations of the tokens prior to
          # removing the paused ones.

          # keep_indices: The indices of the argsort of the auxiliary decoders' 
          # entropy.
          for indices, tokens in zip(keep_indices[::-1], paused_tokens[::-1]):
              tokens[:, indices] = X_L
              X_L = tokens

          return X_L
    \end{minted}
  \vspace{-10pt}
\end{algorithm}



\section{Backbone details}\label{app:sec:backbonedeets} 

In this paper, we use two transformer backbones: ViT-Ti(ny) and ViT-S(mall). We do not experiment with ViT-B, ViT-L
architectures, due to our computational resource constraints. In \cref{app:tab:backbonedeets}, we describe the main
architecture details of the ViT backbones.

\begin{table}[h]
  \centering
  \begin{tabular}{@{}lcccc@{}}
    \toprule
    Model Name & Layers & Embedding dim & Heads & Params \\ \midrule
    ViT-Ti     & 12     & 192           & 3     & 5.9M  \\
    ViT-S      & 12     & 384           & 6     & 22.5M \\\bottomrule
  \end{tabular}
  \vspace{2ex}
  \caption{ViT architectural details used}
  \label{app:tab:backbonedeets}
\end{table}



\section{Brief introduction to Segmenter}\label{app:sec:method:segmenter}

Segmenter\citep{strudel2021segmenter} network ingests an input image $I \in \R^{W\times H\times 3}$ and assigns one of
the $K$ output classes to each input pixel.
From $I$, the model first extracts non-overlapping patches of size $P$, creating a total of $N = \frac{WH}{P^2}$ patches
(also called tokens). Each of those patches are then transformed using a linear embedding layer $E: \R^{3
P^2}\rightarrow \R^{D}$, giving a feature representation $X_0\in \R^{N\times D}$, as shown in \cref{fig:overall_method}.

This feature representation is refined by processing through $L$ transformer encoder layers $T_l$ ($l\in [L]$), where
each transformer encoder layer consists of a multi-head self-attention (MHSA) block followed by a two layers perceptron
(FFN). The overall operation can be represented as:
\[X_L \coloneqq T_L \circ T_{L-1} \circ \dots \circ T_1 \circ E(I) \in \R^{N\times D}\]
Each $T_i$ is residual in nature, \ie $T_i(x) = x + A(x)$ where $A$ encompasses the self-attention and the
multi layer perceptron.

After $L$ layers of such processing, the refined features $X_L$ are fed into a decoder $M_D$. The paper investigated two
kinds of decoders: (a) a linear decoder that takes in the features in $X_L \in \R^{N\times D}$ and produces logits
$\in\R^{N\times K}$ using a $1\times 1$ convolution, (b) a mask transformer which learns $K$ class embeddings that are
jointly processed with $X_L$ through several transformer encoder layers (\`a la $T_i$), and produces a logits
$\in\R^{N\times K}$ as a dot product between the features and the learned class embeddings. The output of either decoder
is reshaped to $\R^{\frac{W}{P}\times \frac{H}{P}\times K}$, and then bilinearly upsampled to produce a logit map of
size $W\times H\times K$. A softmax layer is used to obtain a categorical distribution over the labels for every pixel.
All the layers, $E$, $T_i$, $M_D$ are trained using the standard cross entropy loss.

\section{Patch pausing's limitations}\label{app:sec:pausinglimits}
\begin{figure}[htpb]
  \centering
  \includegraphics[width=0.9\linewidth]{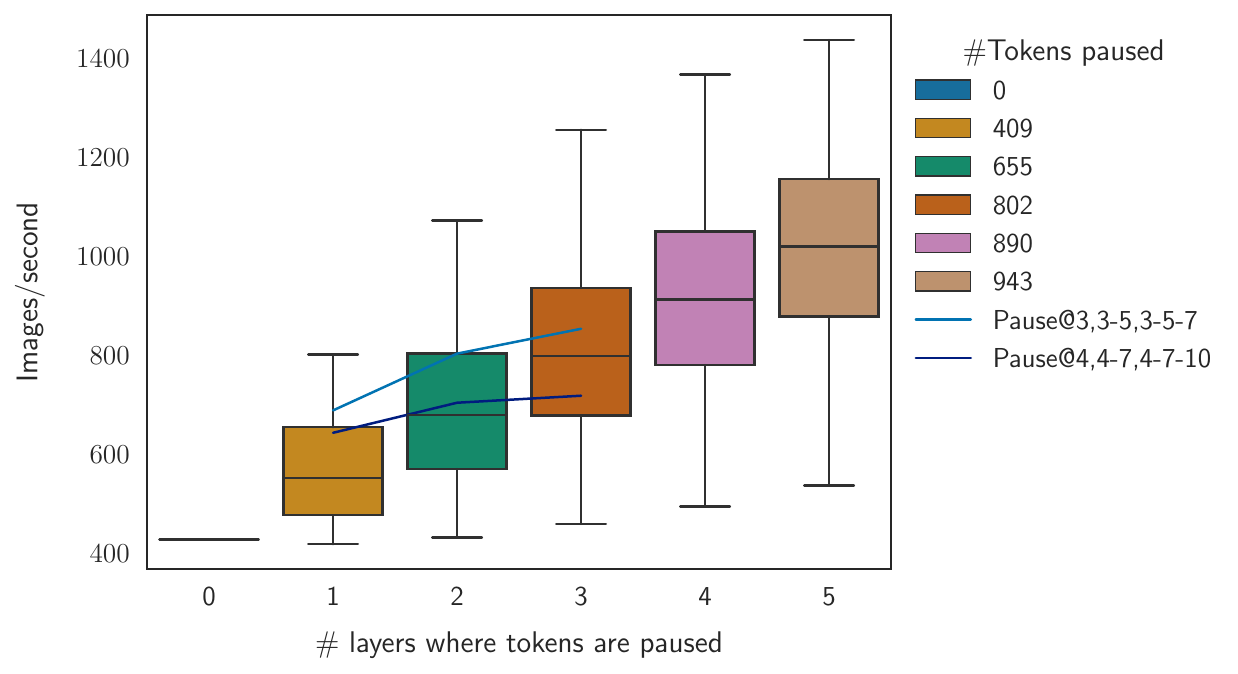}
  \vspace{1ex}
  \caption{Evolution of the throughput with the number of layers using patch pausing.}
  \label{app:fig:layervsimgsec}
\end{figure}
In the main paper, we investigated pausing up to three times in the network. It is indeed tempting to pause at more
layers, but this entails two additional costs: first to compute the entropy and rank the patches, and then the patch
assembly as detailed in \cref{app:sec:assemblerps}. These costs are (ideally) off-set by the reduction in number of patches processed.

To illustrate this further, let us take the case of our experiment with ViT-T backbone (with $12$ layers).
In this experiment, we are interested in how the pausing patterns affect the throughput computed in images processed per second.
To simplify the analysis, we assume that we pause a fixed proportion $\pauseprop = 0.4$.

In \cref{app:fig:layervsimgsec}, we show the distribution of throughput (in images/sec) vs the number of layers we pause tokens at. We use a box-plot where horizontal lines indicates quartiles.
For each value $k$ on the $x$-axis, there are $\binom{12}{k}$
pause configurations.

Consider the case of pausing once. In this case, not all configurations of pausing are useful; pausing too late may in fact be slower than the baseline of not pausing at all, as it incurs an additional cost of auxiliary decoding and patch re-assembling that may offset the time gain of not processing some patches.
This trend is visible on \cref{app:fig:layervsimgsec} and holds even as the number of layers to pause at increases.

We now focus on two cases of pausing: pausing after layers 3, (3, 5), (3, 5, 7) and 4, (4, 7), (4, 7, 10). This two configurations are plotted as lines on \cref{app:fig:layervsimgsec}. 
We see clearly that pausing more has benefits in number of images processed, but that this benefit can quickly plateau if we pause at later layers of the network.
Additionally, this analysis does not consider the mIoU at all.
Indeed, while pausing early-on and at many layers is tempting,
the drop in mIoU becomes too high for those pause configurations to be useful (see \cref{fig:mious_tradeoff_plot,fig:ade_miou_tradeoff}).
Thus the primary limitation is posed by the drop in mIoU rather than throughput.

\section{Influence of the training pause ratio $\pauseprop_l - \pauseprop_h$}

\begin{figure}[htpb]
  \centering
  \includegraphics[width=0.7\linewidth]{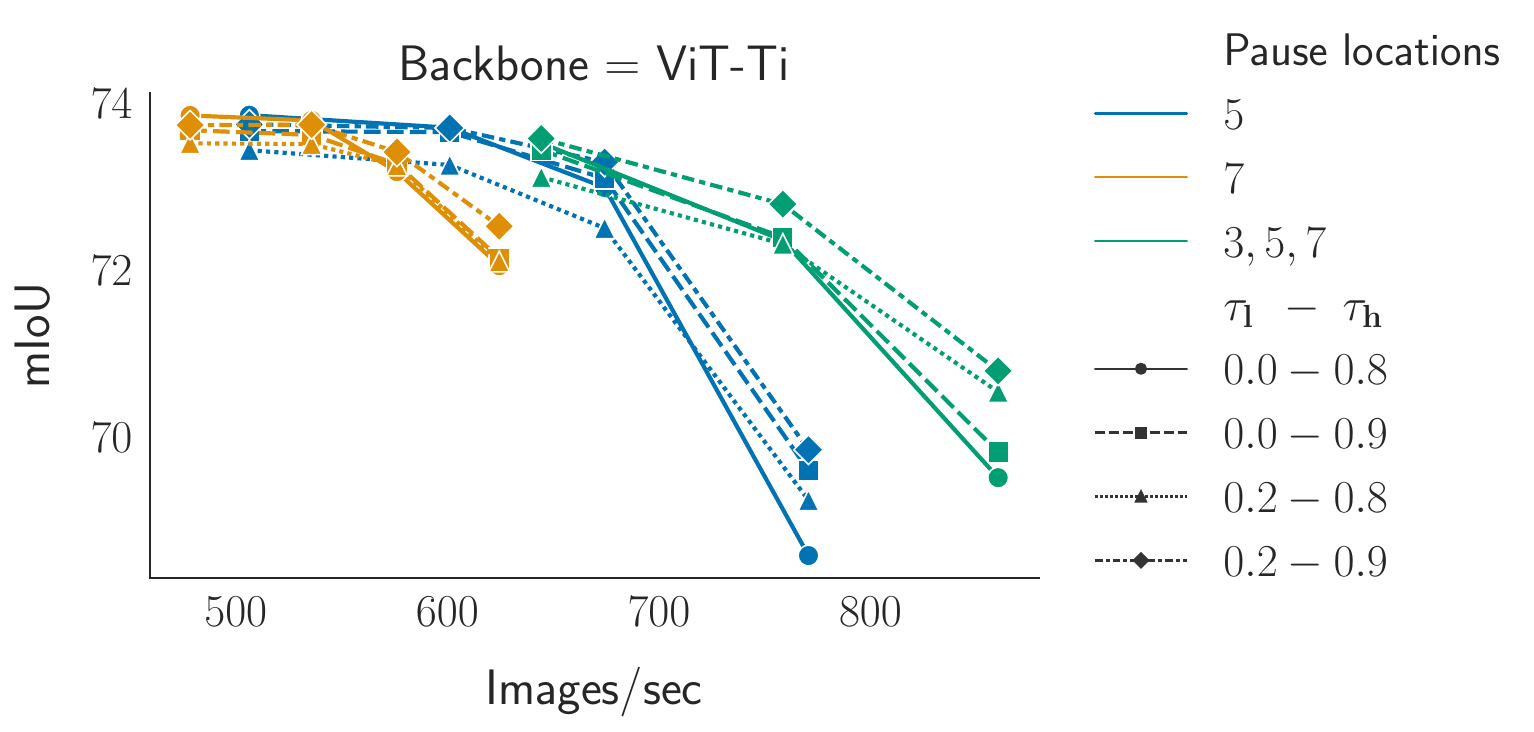}
  \vspace{1ex}
  \caption{mIoU vs throughput for different values of the range of the pause ratio $\pauseprop_l - \pauseprop_h$ introduced in \cref{sec:trainingpaumer}.
  }
  \label{app:fig:pauseratiosweep}
  \vspace{-10pt}
\end{figure}

In \cref{app:fig:pauseratiosweep}, we plot the mIoU of different pause configurations as a function of the throughput for different values of the range of the pause ratio $\pauseprop_l - \pauseprop_h$ introduced in \cref{sec:trainingpaumer}.
We see that the results for various train pause ranges is relatively stable for low amounts of inference pausing ratios. Segmenter's standard deviation (over 3 runs) is $0.35\%$, and we see that the absolute difference in the performance at a given IMPS is about $0.5\%$. This, however, changes when the amount of pausing increases (each colored curve's right corners), when the performance difference is higher ($\approx 1\%$). More aggressive pausing at training seems beneficial ($0-0.9$ performs the best). However, as a middle ground to the multiple configurations investigated, we use $0.2-0.8$ for all our experiments.

\section{Trading off mIoU for higher throughput}
We present results for trading off mIoU for throughput. Specifically, we take all our runs (mIoU vs IMPS) data, and fit
a linear spline, and use the resultant function to predict the mIoU for 8 intermediate IMPS within the range for which
we have experimental results for ViT-Ti in \cref{fig:mious_tradeoff_plot}.
This is to illustrate that we can choose a
pause configuration that fits our run-time requirements (IMPS), and that it works with the performance specified here.

\begin{table}[h]
    \resizebox{\textwidth}{!}{
  
      \begin{tabular}{@{}p{1.5cm}|l|rrrrrrrr@{}}
        \toprule
        Backbone                                 &                     & \multicolumn{1}{l}{} & \multicolumn{1}{l}{} & \multicolumn{1}{l}{} & \multicolumn{1}{l}{} & \multicolumn{1}{l}{} & \multicolumn{1}{l}{} & \multicolumn{1}{l}{} & \multicolumn{1}{l}{} \\ \midrule
        \multirow{5}{1.5cm}{ViT-S \\ 210 im/s}  & {Images / second}     & 252           & 276           & 300           & 325           & 349           & 373           & 397           & 421           \\
        \cmidrule{2-10}
                                                 & {mIoU of \Paumer}   & 77.04            & 76.96            & 76.89            & 76.41            & 76.17            & 74.89            & 73.60            & 71.11            \\
                                                 & {Diff to Segmenter} & -0.03            & -0.12            & -0.19            & -0.66            & -0.90            & -2.18            & -3.47            & -5.96            \\
        \cmidrule(l){2-10}
                                                 & {mIoU of RP}        & 76.68            & 76.08            & 75.77            & 74.79            & 73.32            & 70.75            & 67.60            & 62.64            \\
                                                 & {Diff to Segmenter} & -0.39             & -0.99            & -1.30            & -2.28            & -3.75            & -6.32            & -9.47            & -14.44           \\
        \midrule
        \midrule
        \multirow{5}{1.5cm}{ViT-Ti\\ 424 im/s} & {Images / second}     & 508           & 557           & 605           & 654           & 702           & 751           & 799           & 847           \\
        \cmidrule{2-10}
                                                 & {mIoU of \Paumer}   & 73.42            & 73.35            & 73.23            & 72.91            & 72.76            & 72.37            & 70.99            & 70.58            \\
                                                 & {Diff to Segmenter} & -0.42            & -0.50            & -0.61            & -0.94            & -1.09            & -1.48            & -2.86            & -3.27            \\
        \cmidrule(l){2-10}
                                                 & {mIoU of RP}        & 72.59            & 71.96             & 71.35            & 70.64            & 69.56            & 68.66            & 65.07            & 64.98            \\
                                                 & {Diff to Segmenter} & -1.26            & -1.89            & -2.50            & -3.20            & -4.28            & -5.19            & -8.78             & -8.87             \\
        \bottomrule
      \end{tabular}
    }
    \vspace{2ex}
    \caption{Trading off mIoU for speed. In \cref{sec:exps:subsec:perf}, we showed the performance of mIoU for $50\%$, and
      $100\%$ increase in IMPS. Here we show numbers for a finer grid of IMPS, up to doubling of IMPS.}
    \label{app:tab:mious_imps}
  \end{table}

\section{Influence of the auxiliary loss weight $\lambda$}

\begin{figure}[htpb]
  \centering
  \includegraphics[width=0.7\linewidth]{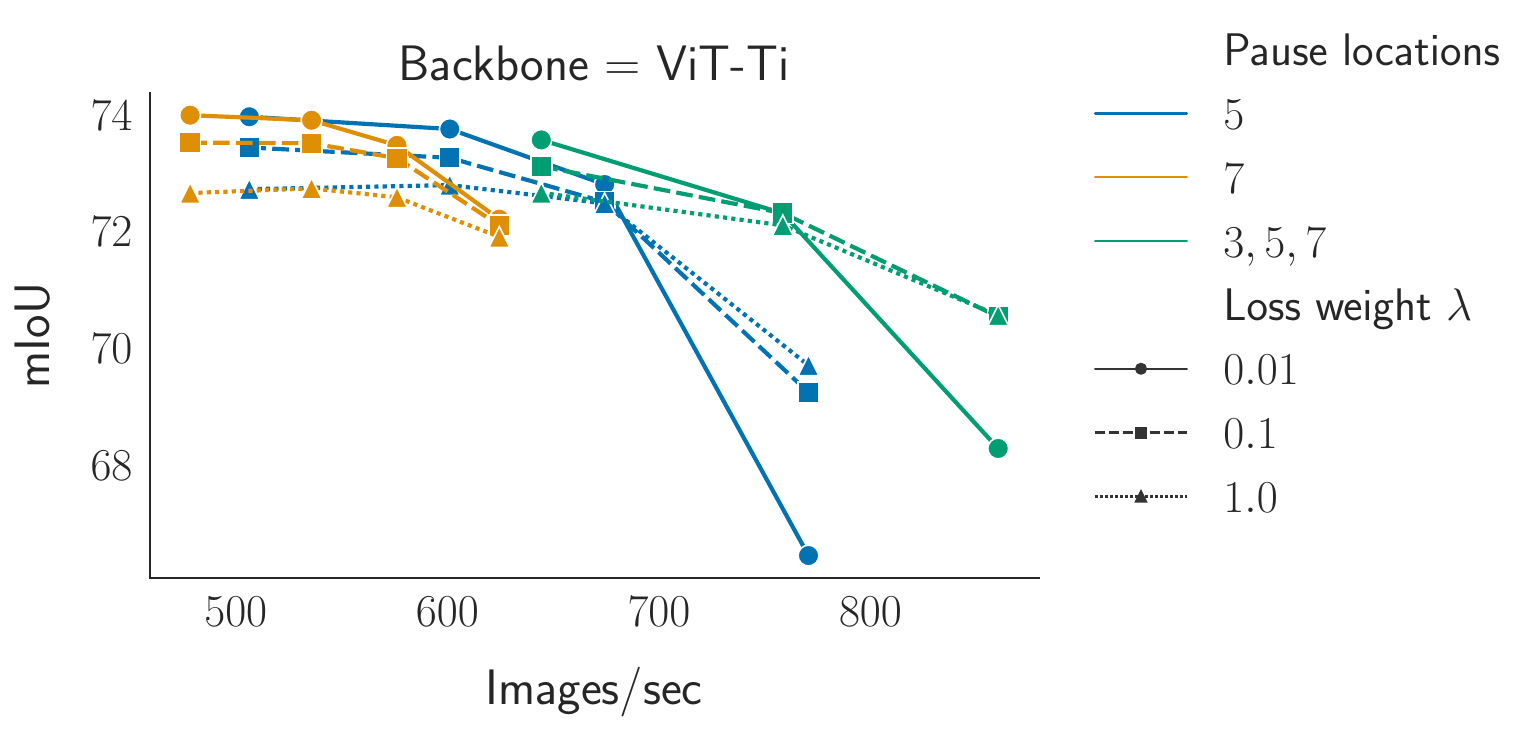}
  \vspace{2ex}
\caption{mIoU vs throughput for different values of the auxiliary loss weight $\lambda$ introduced in
\cref{sec:trainingpaumer}. Lower values of $\lambda$ lead to better performance when pausing few tokens but worse
performance when pausing more, and conversely for higher values of $\lambda$. Our chosen value of $\lambda=0.1$ is a
trade-off that can also be modified depending on the use-case.  
  }
  \label{app:fig:lossweight}
\end{figure}

In \cref{app:fig:lossweight}, we plot the mIoU of different pause configurations as a function of the throughput for
different values of the auxiliary loss weight $\lambda$ introduced in \cref{sec:trainingpaumer}. We can see that
increasing $\lambda$ pushes the network to be more robust to token-pausing but leads to lower performance when pausing
fewer tokens. Thus, $\lambda$ can be tuned depending on the use-case to favor either pausing lesser or a larger number
of tokens.

\section{Interplay of Pause location and Pausing proportion \pauseprop}

\begin{figure}[H]
  \centering
  \includegraphics[width=0.7\linewidth, keepaspectratio]{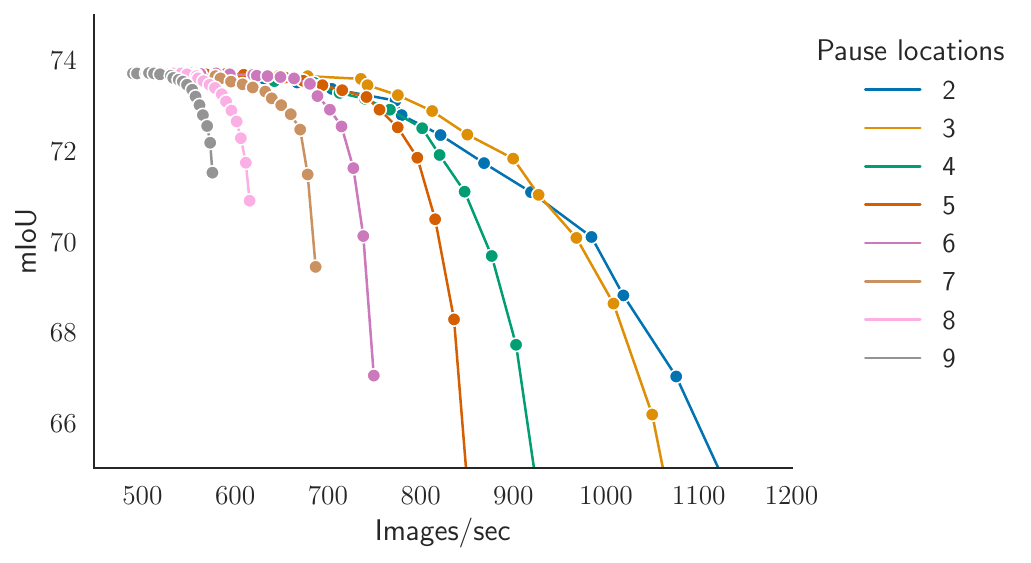}
  \vspace{10pt}
  \caption{Pausing once at various layers for various pause proportions for ViT-Ti on Cityscapes. We see that the highest gains in IMPS are achieved by dropping in earlier layers.}
  \label{app:fig:droponce}
\end{figure}

We showed the results for some configurations in \cref{table:exp_configs}. In this section, we study the interplay
between pause location and pause proportion \pauseprop. In \cref{app:fig:droponce}, we show a sweep over pause
configrations. For each layer, we choose 20 pause proportions in $(0, 1)$. It is apparent that dropping at a later layer
results in lesser drop in mIoU but also does not result in a large gain in IMPS. Thus depending on the desired run-time,
one can choose a pause location and pause proportion that gives the required performance.

\begin{figure}[H]
  \centering
  \includegraphics[width=0.8\linewidth, keepaspectratio]{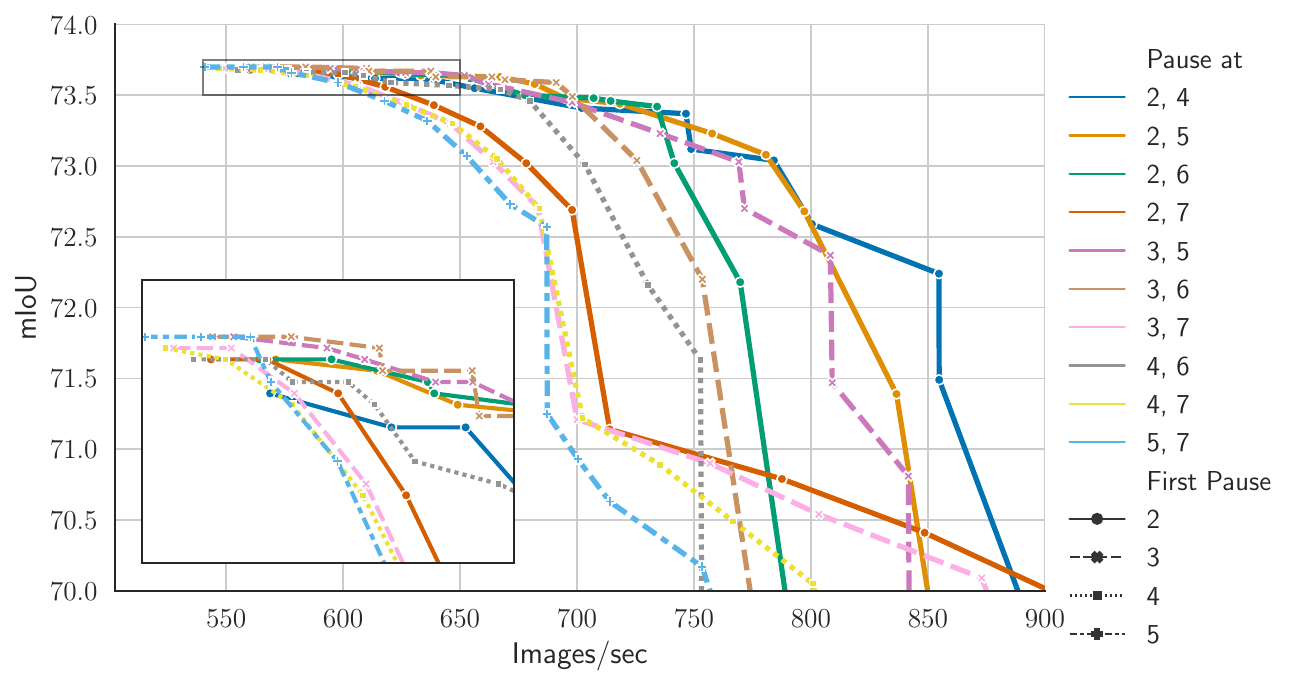}
  \vspace{2ex}
  \caption{Pareto front of pausing at two layers for ViT-Ti on Cityscapes.}
  \label{app:fig:paretofronttwice}
  \vspace{-10pt}
\end{figure}

To examine this further, we plot the performance of pausing twice in \cref{app:fig:paretofronttwice}. Similarly to
the case of pausing once, here we sweep over 10 thresholds for each location, thereby generating 100 configurations for a
given tuple of layers. For those 100 configurations, we plot the pareto front of performance in
\cref{app:fig:paretofronttwice}. We focus on the first pausing layer also, as it has a larger influence on the IMPS gain. 
We can see that pausing at earlier layers leads to a higher increase in IMPS, that pausing small proportions at these layers leads to
slightly higher drop in mIoU than pausing a higher amount in later layers. 

\section{Using Early Exit at test time}\label{app:sec:earlyexit}

\begin{figure}[!h]
    \centering
    \includegraphics[width=0.9\linewidth]{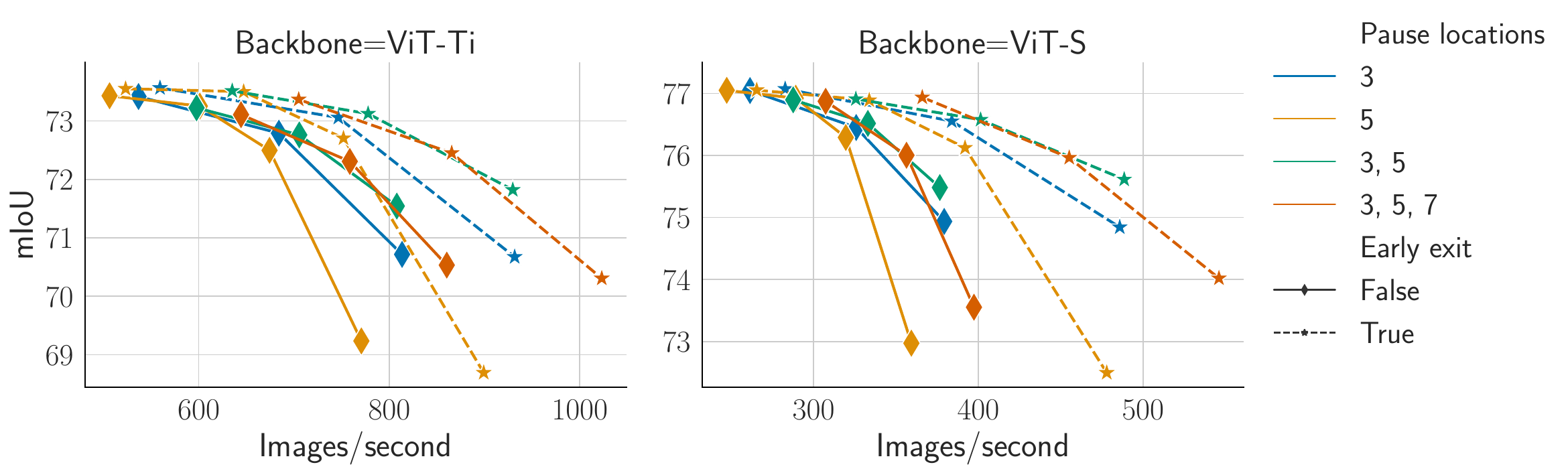}
    \includegraphics[width=0.9\linewidth]{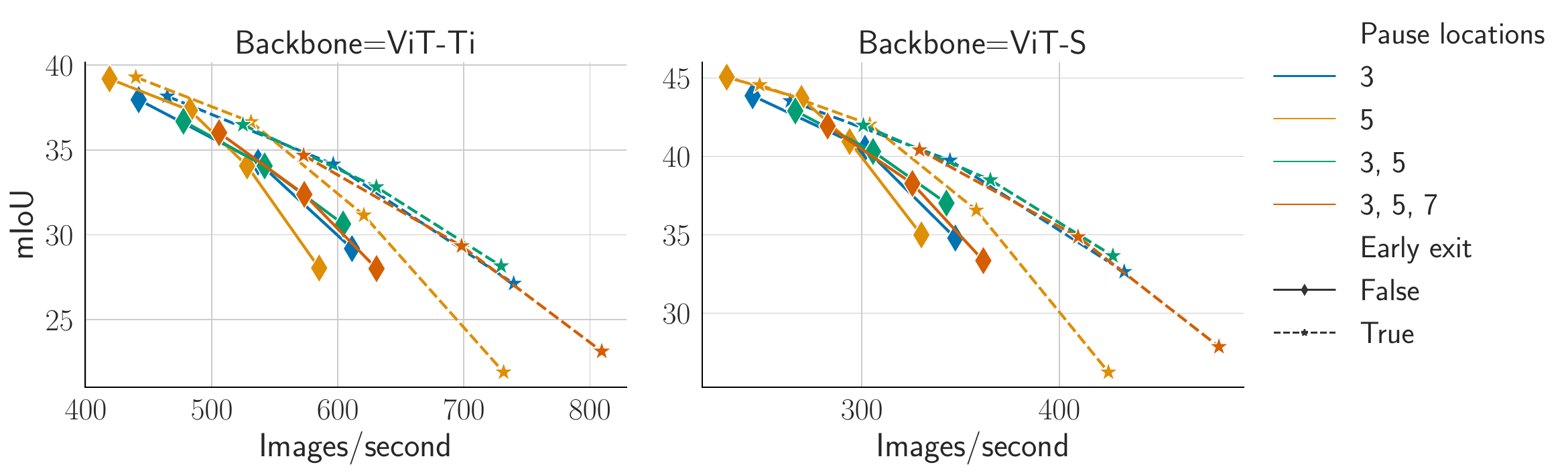}
  \vspace{2ex}
  \caption{Comparison when segmenting with and without early exit at test time on CityScapes and ADE20K.}
  \label{app:fig:earlyexit}
\end{figure}

In \cref{app:fig:earlyexit}, we study the use of early exit on a trained \paumer. Early exit (see works in
\cref{sec:prevwork}) refers to stopping processing of an input once it is deemed to have been processed enough. In our
method, we pause tokens, \ie we stop processing a token by the encoder, and feed it to the decoder to predict the
segmentation label. Here, we compare it with directly using the predictions of the auxiliary decoder itself, without
stopped tokens being processed by the main decoder. \Paumer\, can be run with or without early-exit depending on the
task, and using early exit on a trained Segmenter is straightforward as it does not need any retraining due to the use
of auxiliary decoders. For the same pausing configurations as in \cref{fig:mious_tradeoff_plot}, a network with early
exit runs at a higher throughput with less FLOPs by design, but it may run with a lower mIoU. On
\cref{app:fig:earlyexit}, we see that for Cityscapes, it is beneficial to use \paumer\,with early exit. This finding
might not hold in general, as a complex mask decoder maybe needed for different datasets.

\section{Comparing SETR, Segmenter, EarlyExit using Segmenter}

\begin{figure}[!h]
    \centering
    \includegraphics[width=0.8\textwidth]{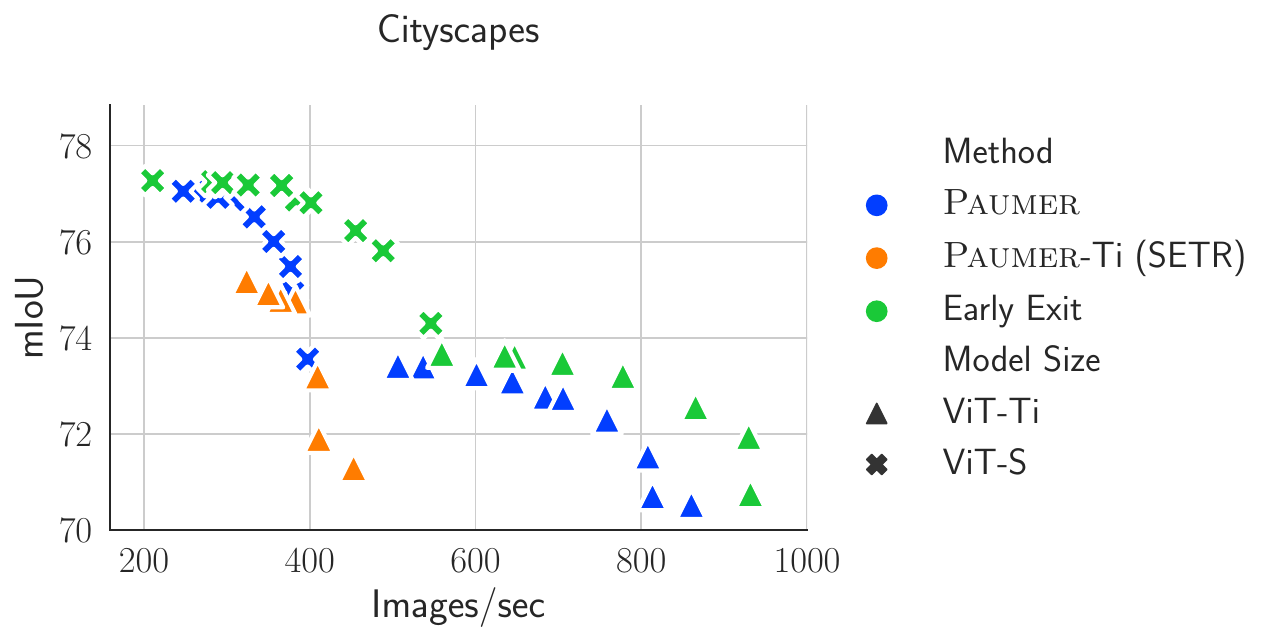}
    \vspace{2ex}
    \caption{We compare \paumer\, on Segmenter, on SETR, and using an early exit variant of \paumer\, on Segmenter. 
Early exit fares better than \paumer\, on Cityscapes. This may be attributable to the dataset, as Cityscapes might not
need more complex mask decoder for accuracy.}
\end{figure}
In \cref{app:sec:earlyexit} we showed the results of early exit. Here we examine the best performances obtained for each
throughput across pause configurations. We estimate this by computing the skyline queries. We see that early exit
performs consistently better on Cityscapes.

Additionally, we implement our patch pausing strategy, \paumer, on the network architecture
SETR\citep{zheng2021rethinking}. SETR's performance drops off more rapidly than Segmenter based patch pausing.
Note that we adapt SETR's PUP decoder to use it with a ViT-Tiny backbone. In particular, we reduce the number of channels of
the decoder to 192, the number of convolutions in the decoder from 4 to 2 and the upscale factor from 2 to 4.

\section{Importance of task specific pretraining}

\begin{figure}[!h]
    \centering
        \centering
        \includegraphics[width=0.7\linewidth]{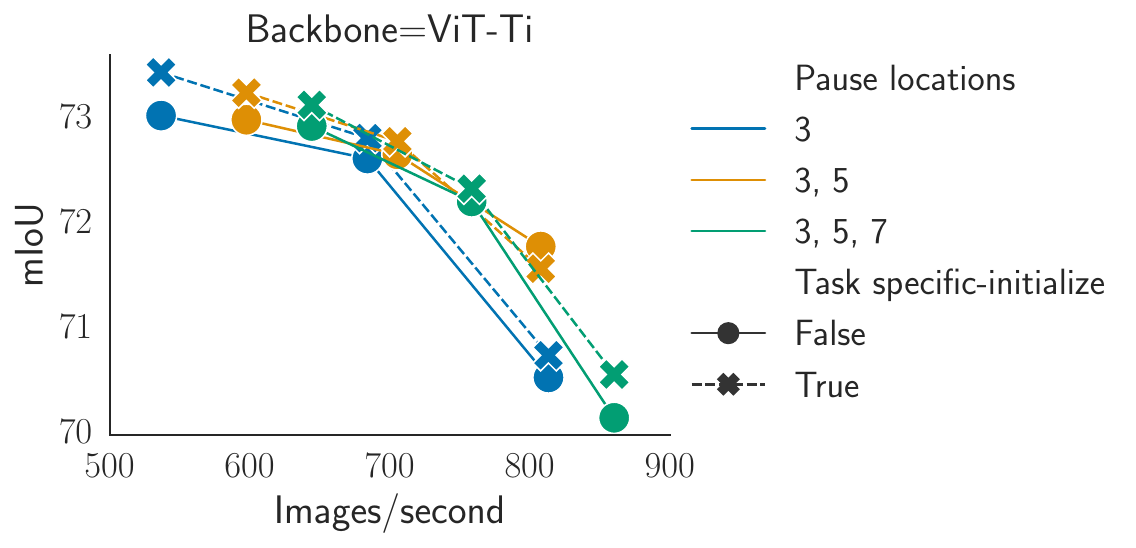}
        \vspace{1ex}
        \caption{Importance of initialization for \Paumer. Task specific initialization benefits performances. We train ViT-Ti
            \paumer: solid lines are from ImageNet pretrained backbones, dashed lines are from Cityscapes pretrained Segmenters.
            Each marker is a configuration from \cref{table:exp_configs}.
        }\label{fig:dropincomparison}
\end{figure}
In \cref{fig:dropincomparison}, we study importance of initialization. We compare ViT-Ti pretrained on Cityscapes (task
specific), and pretrained on ImageNet (generic) and study their impact on performance. When using a generic pretrained
model, our training with \paumer\xspace is increased to 160K iterations instead of 80K. It is apparent that using a task
specific pretrained model brings a relatively consistent benefit in this context.

\section{Entropy as a measure of patch-pausing}
\begin{figure}[h]
  \includegraphics[width=\linewidth,keepaspectratio]{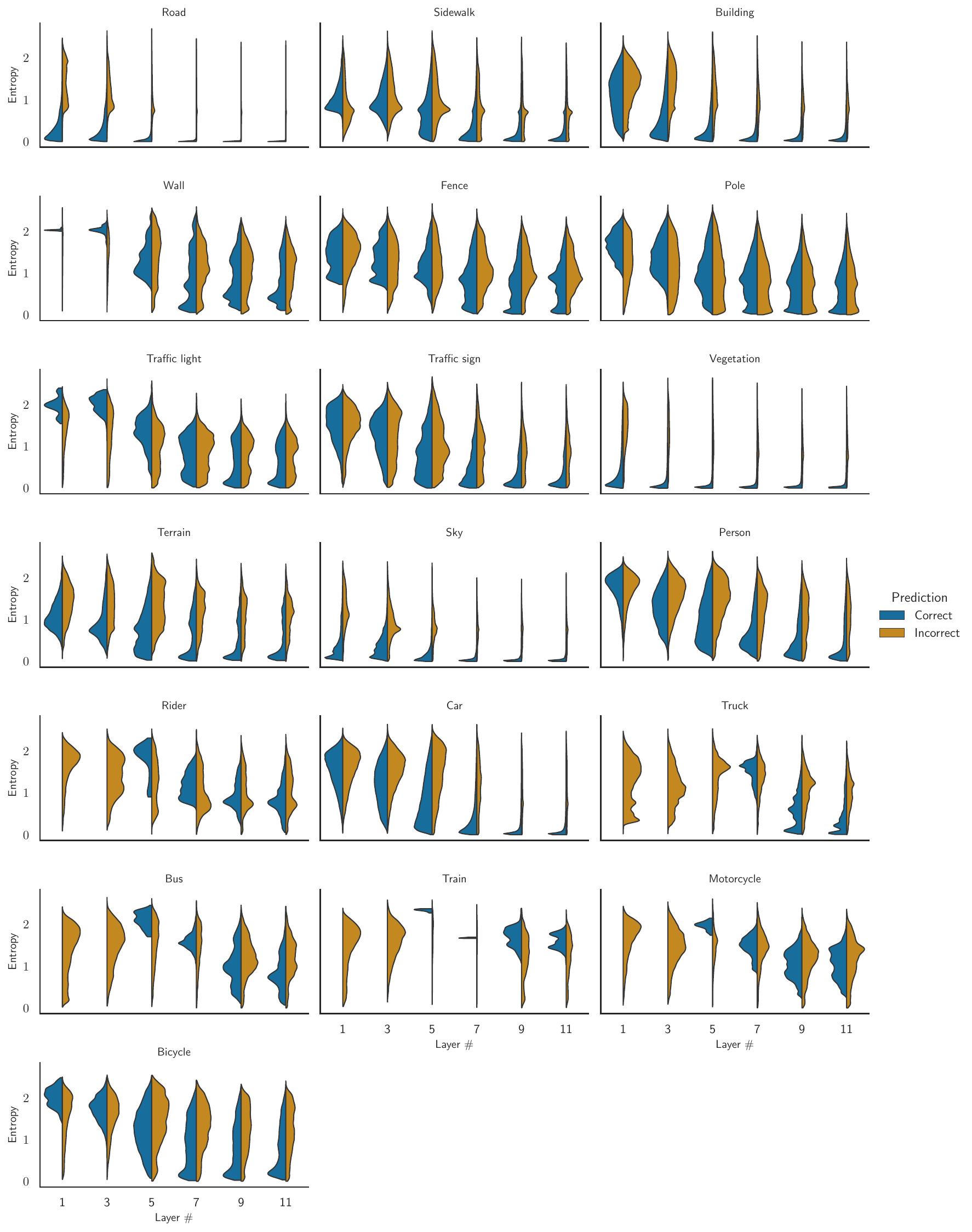}
  \caption{Entropy per layer for each class of Cityscapes. Continues \cref{fig:ent_perf}.}\label{app:fig:ent_perf_per_class}
\end{figure}

In \cref{sec:whyentropy}, we argued that entropy is a reasonable indicator of completion of processing. For that, we
used the illustration in \cref{fig:ent_perf} to show the increase in separation of entropy histograms for pixels
predicted correctly and incorrectly. We expand that in \cref{app:fig:ent_perf_per_class}, to analysing that to each class
individually. The larger separation in entropy in the first few layers of network in prevalent in large classes like
road, building, vegetation, car. As seen in \cref{fig:method_illustration} too these larger classes are paused to gain
IMPS. Smaller, rarer classes like train, motorcycle, rider are tougher to learn and are unlikely to be paused (as evidenced by
their higher entropy).

